\documentclass[10pt,twocolumn,letterpaper]{article}

\usepackage{cvpr}
\usepackage{xcolor, soul}
\usepackage{times}
\usepackage{epsfig}
\usepackage{graphicx}
\usepackage{amsmath}
\usepackage{amssymb}
\usepackage{csquotes}
\usepackage{algorithm}
\usepackage{algorithmic}
\usepackage[british]{babel}
\usepackage{multirow}
\usepackage{amsmath,bm}
\usepackage{mysymbols}
\usepackage{booktabs}       % professional-quality tables
\usepackage{tabularx}
\usepackage{caption}
\usepackage{bbm, dsfont}
\usepackage{enumitem}
\setul{}{1.5pt}
\usepackage{float}
\usepackage{hyperref}

\usepackage[author={Devi, Richard, Caiming, Jiasen}]{pdfcomment}

 \cvprfinalcopy % *** Uncomment this line for the final submission

\newcommand \rot[1] {\rotatebox{75}{#1}}

\ifcvprfinal\fi
\begin{document}

%%%%%%%%% TITLE
\title{
Knowing When to Look: Adaptive Attention via \\A Visual Sentinel for Image Captioning}

\author{Jiasen Lu$^2$\thanks{The major part of this work was done while J. Lu was an intern at Salesforce Research.} \thanks{Equal contribution}, Caiming Xiong$^1$\footnotemark[2], Devi Parikh$^3$, Richard Socher$^1$ \\
$^1$Salesforce Research, $^2$Virginia Tech, $^3$Georgia Institute of Technology\\
{\tt\small jiasenlu@vt.edu, parikh@gatech.edu, \{cxiong, rsocher\}@salesforce.com}
}

\maketitle
%\thispagestyle{empty}
%%%%%%%%% ABSTRACT
\begin{abstract}
Attention-based neural encoder-decoder frameworks have been widely adopted for image captioning. Most methods force visual attention to be active for every generated word. However, the decoder likely requires little to no visual information from the image to predict non-visual words such as ``the'' and ``of''. Other words that may seem visual can often be predicted reliably just from the language model e.g., ``sign'' after ``behind a red stop'' or ``phone'' following ``talking on a cell''. In this paper, we propose a novel adaptive attention model with a visual sentinel. At each time step, our model decides whether to attend to the image (and if so, to which regions) or to the visual sentinel. The model decides whether to attend to the image and where, in order to extract meaningful information for sequential word generation. We test our method on the COCO image captioning 2015 challenge dataset and Flickr30K. Our approach sets the new state-of-the-art by a significant margin. The source code can be downloaded from \href{url}{\textcolor{blue}{https://github.com/jiasenlu/AdaptiveAttention}}

\end{abstract}
%%%%%%%%% BODY TEXT
%%%%%%%%%%%%%%%%%%%%%%%%%%%%%%%%%%%%%%%%%%%%%%%%%%%%%%%%%%%
\vspace{-3mm}
\section{Introduction}
\label{sec:intro}
%%%%%%%%%%%%%%%%%%%%%%%%%%%%%%%%%%%%%%%%%%%%%%%%%%%%%%%%%%%
Automatically generating captions for images has emerged as a prominent interdisciplinary  research problem in both academia and industry. \cite{ fang2015captions, karpathy2015deep, mao2014deep, socher2014grounded, vinyals2015show, Xu2015show}. 
It can aid visually impaired users, and make it easy for users to organize and navigate through large amounts of typically unstructured visual data. In order to generate high quality captions, the model needs to incorporate fine-grained visual clues from the image. Recently, visual attention-based neural encoder-decoder models \cite{Xu2015show, karpathy2015deep, yang2016encode} have been explored, where the attention mechanism typically produces a spatial map highlighting image regions relevant to each generated word.
	
\begin{figure}[t]
 \centering 
 \includegraphics[width=1\linewidth, scale=1, trim={0in 0.4in 0in 0.4in}, clip]{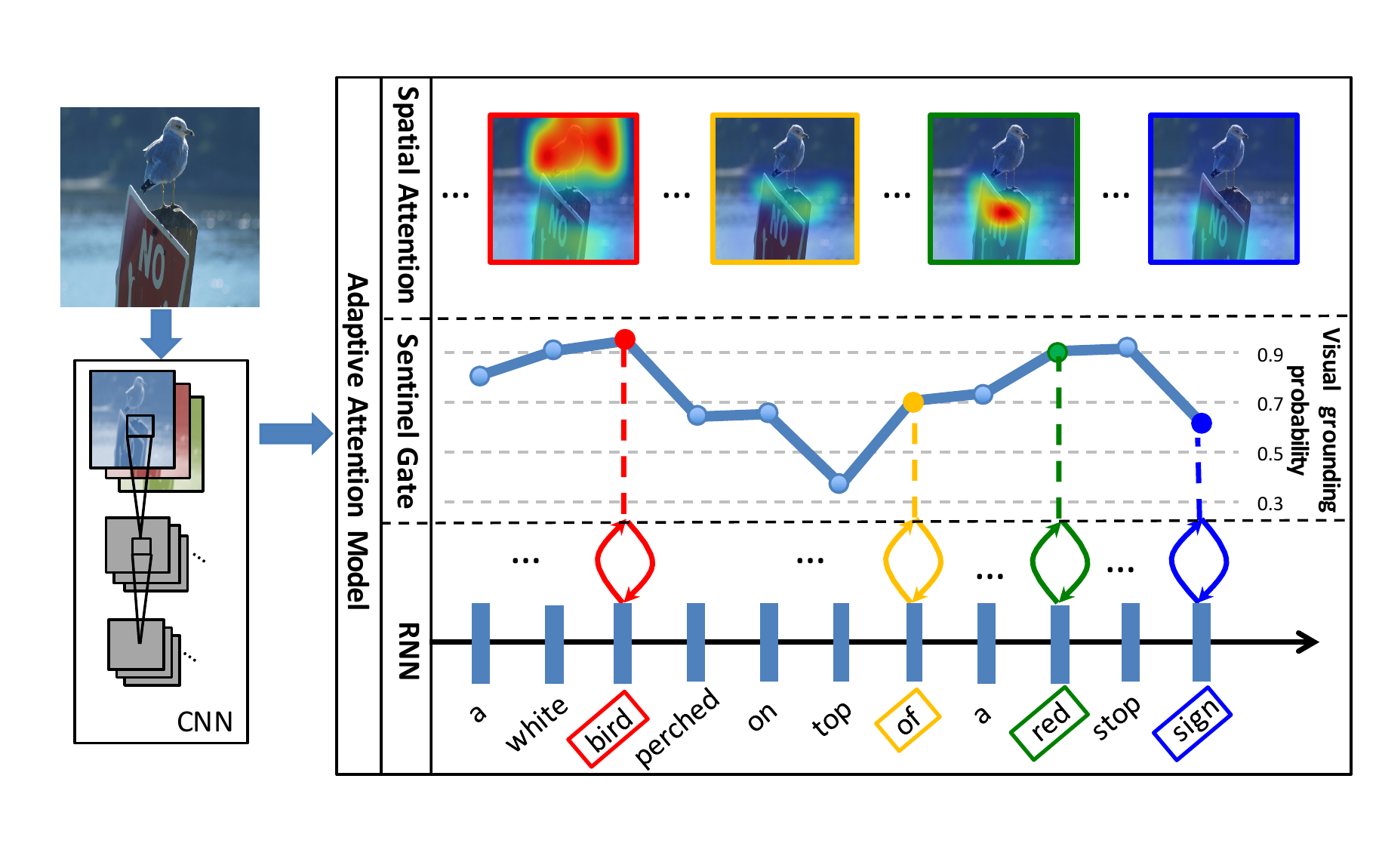}
 \caption{Our model learns an adaptive attention model that automatically determines when to look (\textbf{sentinel gate}) and where to look (\textbf{spatial attention}) for word generation, which are explained in section \ref{sec:spatial}, \ref{sec:adaptive} \& \ref{exp:aaa}.}
\label{fig:teaser}
\vspace{-3mm}
\end{figure}

Most attention models for image captioning and visual question answering attend to the image at every time step, irrespective of which word is going to be emitted next \cite{yang2015stacked,xiong2016dynamic,lu2016hierarchical}.  
However, not all words in the caption have  corresponding visual signals. 
Consider the example in Fig.~\ref{fig:teaser} that shows an image and its generated caption ``A white bird perched on top of a red stop sign''. The words ``a'' and ``of'' do not have corresponding canonical visual signals. Moreover, language correlations make the visual signal unnecessary when generating words like ``on'' and ``top'' following ``perched'', and ``sign'' following ``a red stop''.
In fact, gradients from non-visual words could mislead and diminish the overall effectiveness of the visual signal in guiding the caption generation process.

In this paper, we introduce an adaptive attention encoder-decoder framework which can automatically decide when to rely on visual signals and when to just rely on the language model. 
Of course, when relying on visual signals, the model also decides where -- which image region -- it should attend to. 
%We first propose a new spatial attention model. 
We first propose a novel spatial attention model for extracting spatial image features. Then as our proposed adaptive attention mechanism, we introduce a new Long Short Term Memory (LSTM) extension, which produces an additional ``\textbf{visual sentinel}'' vector instead of a single hidden state.
The ``visual sentinel'', an additional latent representation of the decoder's memory, provides a fallback option to the decoder.
We further design a new sentinel gate, which decides how much new information the decoder wants to get from the image as opposed to relying on the visual sentinel when generating the next word. For example, as illustrated in Fig.~\ref{fig:teaser}, our model learns to attend to the image more when generating words ``white'', ``bird'', ``red'' and ``stop'', and relies more on the visual sentinel when generating words ``top'', ``of'' and ``sign''.

Overall, the main contributions of this paper are:
\begin{itemize}[nolistsep]
\item We introduce an adaptive encoder-decoder framework that automatically decides when to look at the image and when to rely on the language model to generate the next word.
\item We first propose a new spatial attention model, and then build on it to design our novel adaptive attention model with ``visual sentinel''.
\item Our model significantly outperforms other state-of-the-art methods on COCO and Flickr30k.
\item We perform an extensive analysis of our adaptive attention model, including visual grounding probabilities of words and weakly supervised localization of generated attention maps.
\end{itemize}

\section{Method}
\label{sec:method}
%To ease understanding, we first describe the encoder-decoder for image captioning in Sec.~\ref{sec:encoder-decoder}. The proposed spatial attention model and adaptive attention model are then described in Sec.~\ref{sec:spatial} and Sec.~\ref{sec:adaptive} respectively. Finally, Sec.~\ref{sec:details} shows the implementation details. 
We first describe the generic neural encoder-decoder framework for image captioning in Sec.~\ref{sec:encoder-decoder}, then introduce our proposed attention-based image captioning models in Sec.~\ref{sec:spatial} \& ~\ref{sec:adaptive}.
%Note that in the exposition that follows, we omit the bias term $b$ to avoid notational clutter, but they are included in the implementation.
%As shown in Fig. 2., we decompose our model into 4 parts. 1: encoder-CNN component, 2: decoder-RNN component, 3: Spatial attention component and 4: Adaptive attention component.
%
%\subsection{Notation}
%Our model takes a single raw image $I$ and generates a caption $Y = \left\lbrace y_1,\ldots, y_C \right\rbrace$, where $C$ is the length of the Caption. The image is processed by a convolutional neural network (CNN) to extract spatial image feature $S = \left \lbrace s_1,\ldots, s_N \right \rbrace$ and holistic image feature $v$ for the language model. The weights in different module/layers are denoted with $W$ , with appropriate sub/super-scripts as necessary. In the exposition that follows, we omit the bias term b to avoid notational clutter, but they are included in the model.
\subsection{Encoder-Decoder for Image Captioning}
\label{sec:encoder-decoder}
We start by briefly describing the encoder-decoder image captioning framework \cite{vinyals2015show, Xu2015show}. Given an image and the corresponding caption, the encoder-decoder model directly maximizes the following objective:
\begin{equation}
\bm{\theta}^* = \arg \max_{\bm{\theta}} \sum_{(\bm{I},\bm{y})} \log p(\bm{y}\vert \bm{I}; \bm{\theta})
\end{equation}
where $\bm{\theta}$ are the parameters of the model, $\bm{I}$ is the image, and $\bm{y}=\{y_1,\ldots,y_t\}$ is the corresponding caption. Using the chain rule, the log likelihood of the joint probability distribution can be decomposed into ordered conditionals:
\begin{equation}
\log p(\bm{y}) = \sum_{t=1}^T \log p(y_t \vert y_1,\ldots, y_{t-1}, \bm{I})
\label{eq:2}
\end{equation}
where we drop the dependency on model parameters for convenience.  

In the encoder-decoder framework, with recurrent neural network (RNN), each conditional probability is modeled as:
\begin{equation}
\log p(y_t \vert y_1, \ldots, y_{t-1}, \bm{I}) = f(\bm{h}_{t}, \bm{c}_t)
\label{eq:3}
\end{equation}
where $f$ is a nonlinear function that outputs the probability of $y_t$. $\bm{c}_t$ is the visual context vector at time $t$ extracted from image $\bm{I}$. $\bm{h}_{t}$ is the hidden state of the RNN at time $t$. In this paper, we adopt Long-Short Term Memory (LSTM) instead of a vanilla RNN. The former have demonstrated state-of-the-art performance on a variety of sequence modeling tasks. $\bm{h}_{t}$ is modeled as: 
\begin{equation}
\bm{h}_t = \mathrm{LSTM}(\bm{x}_t, \bm{h}_{t-1}, \bm{m}_{t-1})
\end{equation}
where $\bm{x}_{t}$ is the input vector. $\bm{m}_{t-1}$ is the memory cell vector at time $t-1$. 

Commonly, context vector, $\bm{c}_t$ is an important factor in the neural encoder-decoder framework, which provides visual evidence for caption generation~\cite{mao2014deep, vinyals2015show, Xu2015show, you2016image}. These different ways of modeling the context vector fall into two categories: vanilla encoder-decoder and attention-based encoder-decoder frameworks: 
\begin{itemize}[nolistsep]
\item First, in the vanilla framework, $\bm{c}_t$ is only dependent on the encoder, a Convolutional Neural Network (CNN). The input image $\bm{I}$ is fed into the CNN, which extracts the last fully connected layer as a global image feature \cite{ mao2014deep, vinyals2015show}. Across generated words, the context vector $\bm{c}_t$ keeps constant, and does not depend on the hidden state of the decoder. 
\item Second, in the attention-based framework, $\bm{c}_t$ is dependent on both encoder and decoder. At time $t$, based on the hidden state, the decoder would attend to the specific regions of the image and compute $\bm{c}_t$ using the spatial image features from a convolution layer of a CNN. In \cite{Xu2015show, you2016image}, they show that attention models can significantly improve the performance of image captioning. 
\end{itemize}

To compute the context vector $c_t$, we first propose our spatial attention model in Sec.~\ref{sec:spatial}, then extend the model to an adaptive attention model in Sec.~\ref{sec:adaptive}.
\subsection{Spatial Attention Model}
\label{sec:spatial}
First, we propose a spatial attention model for computing the context vector $\bm{c}_t$ which is defined as:
\begin{equation}
\bm{c}_t=g(\bm{V}, \bm{h}_t)
\end{equation}
where $g$ is the attention function, $\bm{V} = \left[\bm{v}_1,\ldots,\bm{v}_k \right], \bm{v}_i \in \mathcal{R}^d$ is the spatial image features, each of which is a $d$ dimensional representation corresponding to a part of the image. $\bm{h}_t$ is the hidden state of RNN at time $t$.

Given the spatial image feature $\bm{V} \in \mathcal{R}^{d \times k}$ and hidden state $\bm{h}_t \in \mathcal{R}^d$ of the LSTM, we feed them through a single layer neural network followed by a softmax function to generate the attention distribution over the $k$ regions of the image:
\begin{eqnarray}
\bm{z}_t &=& \bm{w}_{h}^T \tanh(\bm{W}_v \bm{V} + (\bm{W}_{g} \bm{h}_t)
\bm{\mathbbm{1}}^T ) 
\label{eq:4}
\\ 
\bm{\alpha}_t &=& \textrm{softmax}(\bm{z}_t) 
\label{eq:5}
\end{eqnarray}
where $\bm{\mathbbm{1}} \in \mathcal{R}^k$ is a vector with all elements set to 1. $\bm{W}_v, \bm{W}_g \in \mathcal{R}^{k \times d}$ and $\bm{w}_{h} \in \mathcal{R}^k$ are parameters to be learnt. $\bm{\alpha} \in \mathcal{R}^k$ is the attention weight over features in $\bm{V}$. Based on the attention distribution, the context vector $\bm{c}_t$ can be obtained by:
\begin{equation}
\bm{c}_t = \sum_{i=1}^k \bm{\alpha}_{ti} \bm{v}_{ti}
\end{equation}
where $\bm{c}_t$ and $\bm{h}_t$ are combined to predict next word $y_{t+1}$ as in Equation~\ref{eq:3}.

\begin{figure}[t]
 \centering 
 \includegraphics[width=1\linewidth, scale=1]{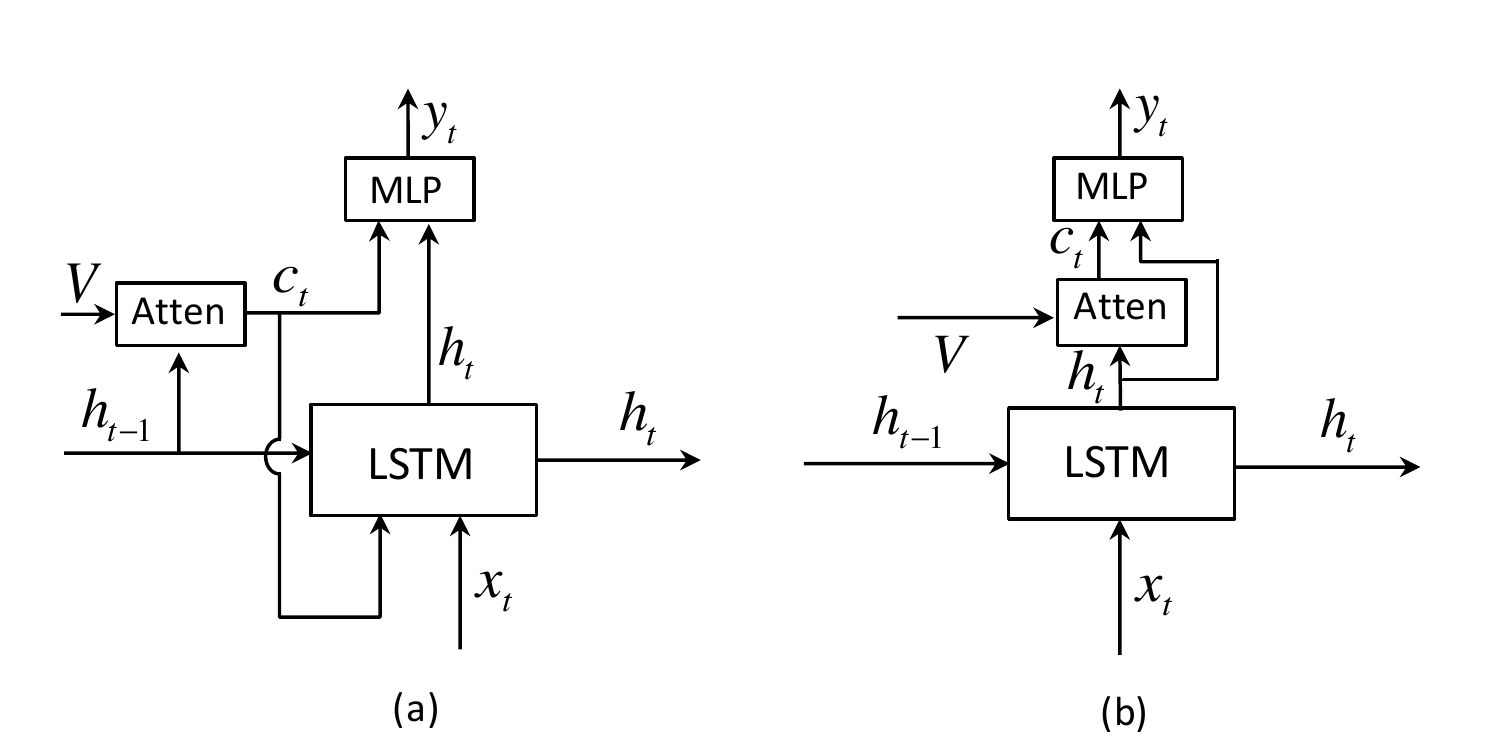}
 \caption{A illustration of soft attention model from \cite{Xu2015show} (a) and our proposed spatial attention model (b).
}
 \label{fig:model_spatial}
 \vspace{-3mm}
\end{figure}

Different from \cite{Xu2015show}, shown in Fig.~\ref{fig:model_spatial}, we use the current hidden state $\bm{h}_t$ to analyze where to look (i.e., generating the context vector $\bm{c}_t$), then combine both sources of information to predict the next word.
Our motivation stems from the superior performance of residual network \cite{he2015deep}. The generated context vector $\bm{c}_t$ could be considered as the residual visual information of current hidden state $\bm{h}_t$, which diminishes the uncertainty or complements the informativeness of the current hidden state for next word prediction. We also empirically find our spatial attention model performs better, as illustrated in Table~\ref{test_result}.
\subsection{Adaptive Attention Model}
\label{sec:adaptive}
While spatial attention based decoders have proven to be effective for image captioning, they cannot determine when to rely on visual signal and when to rely on the language model. In this section, motivated from Merity \etal~\cite{Merity2016Pointer}, we introduce a new concept -- ``visual sentinel'', which is a latent representation of what the decoder already knows. With the ``visual sentinel'', we extend our spatial attention model, and propose an adaptive model that is able to determine whether it needs to attend the image to predict next word. 

\textbf{What is visual sentinel?} 
The decoder's memory stores both long and short term visual and linguistic information. Our model learns to extract a new component from this that the model can fall back on when it chooses to not attend to the image. This new component is called the visual sentinel. And the gate that decides whether to attend to the image or to the visual sentinel is the sentinel gate.
%Motivated from Merity et al. \cite{Merity2016Pointer}, assume that "visual sentinel" is a latent information in the current memory of decoder. It can be used to decided
When the decoder RNN is an LSTM, we consider those information preserved in its memory cell. 
Therefore, we extend the LSTM to obtain the ``visual sentinel'' vector $\bm{s}_t$ by: 
\begin{eqnarray}
\bm{g}_t &=& \sigma \left( \bm{W}_{x} \bm{x}_t + \bm{W}_{h} \bm{h}_{t-1} \right)\\
\bm{s}_t &=& \bm{g}_t \odot \tanh \left( \bm{m}_ t\right)
\end{eqnarray}
where $\bm{W}_x$ and $\bm{W}_h$ are weight parameters to be learned, $\bm{x}_t$ is the input to the LSTM at time step $t$, and $\bm{g}_t$ is the gate applied on the memory cell $\bm{m}_t$. $\odot$ represents the element-wise product and $\sigma$ is the logistic sigmoid activation. 

\begin{figure}[t]
 \centering 
 \includegraphics[width=1\linewidth, scale=1]{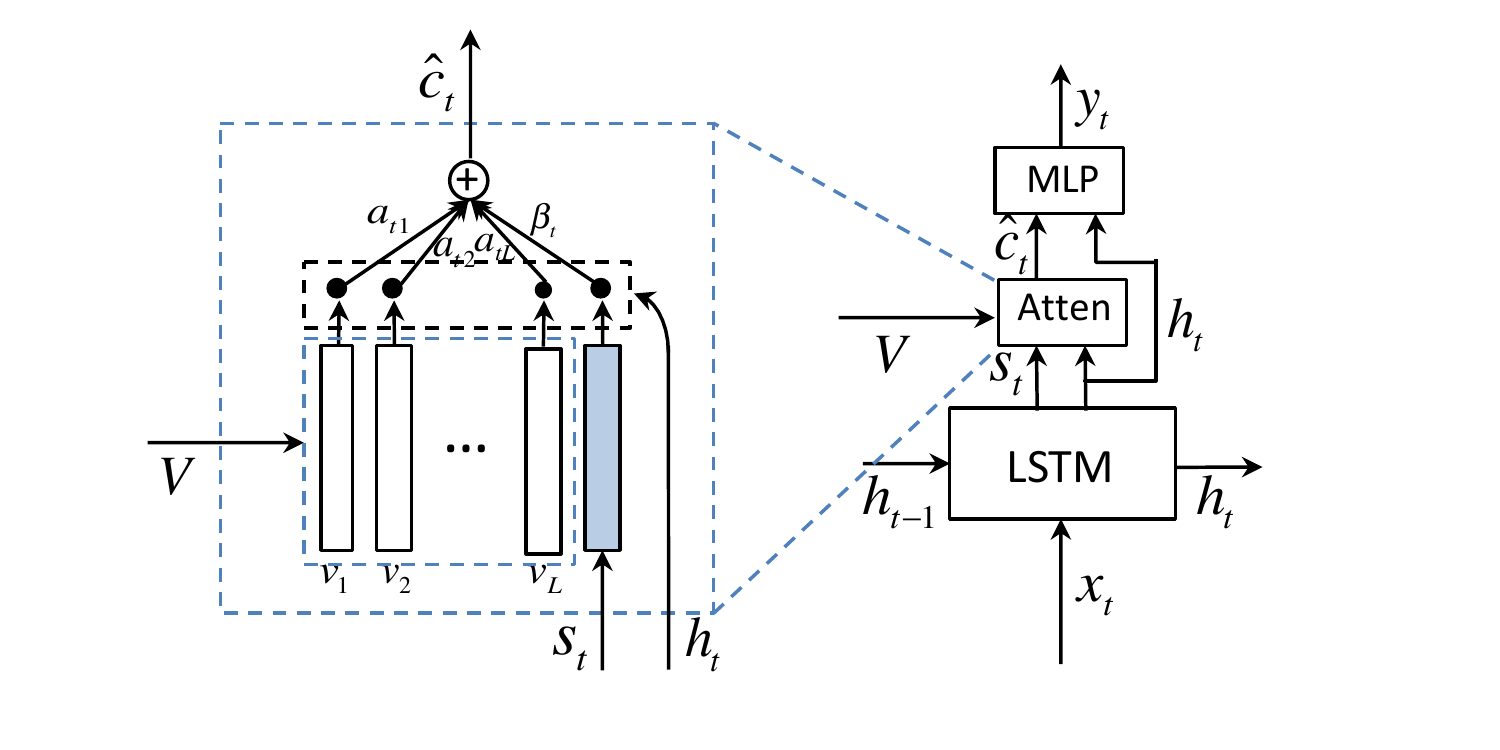}
 \caption{An illustration of the proposed model generating the $t$-th target word $y_t$ given the image.}
 \label{fig:model}
 \vspace{-3mm}
\end{figure}

Based on the visual sentinel, we propose an adaptive attention model to compute the context vector. In our proposed architecture (see Fig.~\ref{fig:model}), our new adaptive context vector is defined as $\hat{\bm{c}}_t$, which is modeled as a mixture of the spatially attended image features (i.e. context vector of spatial attention model) and the visual sentinel vector. This trades off how much new information the network is considering from the image with what it already knows in the decoder memory (i.e., the visual sentinel ). The mixture model is defined as follows:
\begin{equation}
\hat{\bm{c}}_t = \beta_t \bm{s}_t + (1-\beta_t) \bm{c}_t
\end{equation}
where $\beta_t$ is the new sentinel gate at time $t$. In our mixture model, $\beta_t$ produces a scalar in the range $[0, 1]$. A value of $1$ implies that only the visual sentinel information is used and $0$ means only spatial image information is used when generating the next word.

To compute the new sentinel gate $\beta_t$, we modified the spatial attention component. In particular, we add an additional element to $z$, the vector containing attention scores as defined in Equation~\ref{eq:4}. This element indicates how much ``attention'' the network is placing on the sentinel (as opposed to the image features). The addition of this extra element is summarized by converting Equation~\ref{eq:5} to:
\begin{equation}
\hat{\bm{\alpha}}_t = \textrm{softmax}([\bm{z}_t; \bm{w}_{h}^T\tanh(\bm{W}_s \bm{s}_t + (\bm{W}_{g} \bm{h}_t))])
\label{Eq:12}
\end{equation}
where $[\cdot ; \cdot]$ indicates concatenation. $\bm{W}_s$ and $\bm{W}_{g}$ are weight parameters. Notably, $\bm{W}_{g}$ is the same weight parameter as in Equation~\ref{eq:4}. $\hat{\bm{\alpha}}_t \in \mathcal{R}^{k+1}$ is the attention distribution over both the spatial image feature as well as the visual sentinel vector. We interpret the last element of this vector to be the gate value: $\beta_t=\bm{\alpha}_t[k+1]$.

The probability over a vocabulary of possible words at time $t$ can be calculated as:
\begin{equation}
\bm{p}_t = \mathrm{softmax} \left( \bm{W}_p (\hat{\bm{c}}_t + \bm{h}_{t}) \right) 
\end{equation}
where $\bm{W}_p$ is the weight parameters to be learnt.

This formulation encourages the model to adaptively attend to the image vs. the visual sentinel when generating the next word. The sentinel vector is updated at each time step. With this adaptive attention model, we call our framework the adaptive encoder-decoder image captioning framework.%Note that it is crucial for the sentinel vector to come from the decoder RNN. %Simply learning a bias vector as in \cite{Merity2016Pointer} was empirically found to not be sufficient. \devi{Is this true? Did you find this in experiments? Can we give a quick result here or in results?}
%\begin{equation}
%\bm{c}_t=g(\bm{V}, \bm{h}_t, \bm{s}_t)
%\end{equation}
%

\section{Implementation Details}
\label{sec:details}
In this section, we describe the implementation details of our model and how we train our network.

\textbf{Encoder-CNN.} The encoder uses a CNN to get the representation of images. Specifically, the spatial feature outputs of the last convolutional layer of ResNet \cite{he2015deep} are used, which have a dimension of $2048 \times 7 \times 7$. We use $\bm{A} = \{\bm{a}_1, \ldots, \bm{a}_{k}\}, \bm{a}_i \in \mathcal{R}^{2048}$ to represent the spatial CNN features at each of the $k$ grid locations. Following \cite{he2015deep}, the global image feature can be obtained by:
\begin{equation}
\bm{a}^g = \frac{1}{k} \sum_{i=1}^k \bm{a}_i
\end{equation}
where $\bm{a}^g$ is the global image feature.
For modeling convenience, we use a single layer perceptron with rectifier activation function to transform the image feature vector into new vectors with dimension $d$:
\begin{eqnarray}
\bm{v}_i &=& \mathrm{ReLU}(\bm{W}_a \bm{a}_i) \\
\bm{v}^g &=& \mathrm{ReLU}(\bm{W}_b \bm{a}^g)
\end{eqnarray}
where $\bm{W}_a$ and $\bm{W}_g$ are the weight parameters. The transformed spatial image feature form $\bm{V} = \left[\bm{v}_1,\ldots,\bm{v}_k \right]$. 

\textbf{Decoder-RNN.} We concatenate the word embedding vector $\bm{w}_t$ and global image feature vector $\bm{v}^g$ to get the input vector $\bm{x}_t = [\bm{w}_t; \bm{v}^g]$. We use a single layer neural network to transform the visual sentinel vector $\bm{s}_t$ and LSTM output vector $\bm{h}_t$ into new vectors that have the dimension $d$.  

\textbf{Training details.} In our experiments, we use a single layer LSTM with hidden size of 512. We use the Adam optimizer with base learning rate of 5e-4 for the language model and 1e-5 for the CNN. The momentum and weight-decay are 0.8 and 0.999 respectively. We finetune the CNN network after 20 epochs. We set the batch size to be 80 and train for up to 50 epochs with early stopping if the validation CIDEr \cite{vedantam2015cider} score had not improved over the last 6 epochs. Our model can be trained within 30 hours on a single Titan X GPU. We use beam size of 3 when sampling the caption for both  COCO and Flickr30k datasets.
%%%%%%%%%%%%%%%%%%%%%%%%%%%%%%%%%%%%%%%%%%%%%%%%%%%%%%%%%%%
\section{Related Work}
\label{sec:related}
%%%%%%%%%%%%%%%%%%%%%%%%%%%%%%%%%%%%%%%%%%%%%%%%%%%%%%%%%%%
Image captioning has many important applications ranging from helping visually impaired users to human-robot interaction. As a result, many different models have been developed for image captioning. In general, those methods can be divided into two categories: template-based  \cite{farhadi2010every, kulkarni2013babytalk, kuznetsova2012collective,mitchell2012midge} and neural-based  \cite{kiros2014ICML, mao2014deep, devlin2015exploring, chen2015mind, vinyals2015show, donahue2015long, karpathy2015deep, Xu2015show, fang2015captions, you2016image, yang2016encode,yao2016boosting}.

\textbf{Template-based approaches} generate caption templates whose slots are filled in based on outputs of object detection, attribute classification, and scene recognition. Farhadi \etal \cite{farhadi2010every} infer a triplet of scene elements which is converted to text using templates. Kulkarni \etal \cite{kulkarni2013babytalk} adopt a Conditional Random Field (CRF) to jointly reason across objects, attributes, and prepositions before filling the slots. %model to predict labels based on the result of object detection, attribute and preposition recognition, then fills the most likely labels into slots of a template. 
\cite{kuznetsova2012collective,mitchell2012midge} use more powerful  language templates such as a syntactically well-formed tree, and add descriptive information from the output of attribute detection.

\textbf{Neural-based approaches} are inspired by the success of sequence-to-sequence encoder-decoder frameworks in machine translation   \cite{cho2014learning,sutskever2014sequence,bahdanau2014neural} with the view that image captioning is analogous to translating images to text. Kiros \etal \cite{kiros2014ICML} proposed a feed forward neural network with a multimodal log-bilinear model to predict the next word given the image and previous word. Other methods then replaced the feed forward neural network with a recurrent neural network \cite{mao2014deep, chen2015mind}. Vinyals \etal \cite{vinyals2015show} use an LSTM instead of a vanilla RNN as the decoder. However, all these approaches represent the image with the last fully connected layer of a CNN. Karpathy \etal \cite{karpathy2015deep} adopt the result of object detection from R-CNN and output of a bidirectional RNN to learn a joint embedding space for caption ranking and generation. 

\begin{table*}[t]\footnotesize
\center
  \begin{tabular}{l c c c c c c c c c c c c }
    \toprule
    \multicolumn{1}{c}{} & \multicolumn{6}{c}{Flickr30k}  & \multicolumn{6}{c}{MS-COCO}\\
   \cmidrule(r){2-7}
   \cmidrule(r){8-13}
    Method  & B-1 & B-2 & B-3 & B-4 & METEOR & CIDEr & B-1 & B-2 & B-3 & B-4 & METEOR & CIDEr\\
    \midrule
    DeepVS \cite{karpathy2015deep}& 0.573 & 0.369 & 0.240 & 0.157 & 0.153 & 0.247 & 0.625 & 0.450 & 0.321 & 0.230 & 0.195 & 0.660 \\
    Hard-Attention \cite{Xu2015show}& 0.669 & 0.439 & 0.296 & 0.199 & 0.185 & - & 0.718 & 0.504 & 0.357 & 0.250 & 0.230 & - \\
    ATT-FCN$^{\dagger}$ \cite{you2016image}& 0.647 & 0.460 & 0.324 & 0.230 & 0.189 & - & 0.709 & 0.537 & 0.402 & 0.304 & 0.243 & - \\
    ERD \cite{yang2016encode}& - & - & - & - & - & - & - & - & - & 0.298 & 0.240 &0.895 \\	
    MSM$^\dagger$ \cite{yao2016boosting}& - & - & - & - & - & - & 0.730 & 0.565 & 0.429 & 0.325 & 0.251 & 0.986 \\
    \midrule
    Ours-Spatial &0.644&0.462&0.327& 0.231&0.202&0.493& 0.734 & 0.566 & 0.418 & 0.304 & 0.257 & 1.029 \\
	Ours-Adaptive & \textbf{0.677} & \textbf{0.494} & \textbf{0.354} & \textbf{0.251} & \textbf{0.204} & \textbf{0.531} & \textbf{0.742} & \textbf{0.580} & \textbf{0.439} & \textbf{0.332} & \textbf{0.266} & \textbf{1.085} \\
	%Ours-Adaptive$^\dagger$ & \textbf{0.683} & \textbf{0.504} & \textbf{0.365} & \textbf{0.261} & \textbf{0.208} & \textbf{0.547} & \textbf{0.757} & \textbf{0.597} & \textbf{0.459} & \textbf{0.352} &\textbf{0.271} &\textbf{1.133}\\
    \bottomrule
 \end{tabular}
\small{\caption{Performance on Flickr30k and COCO test splits. $\dagger$ indicates ensemble models. \textbf{B-n} is BLEU score that uses up to n-grams. Higher is better in all columns. For future comparisons, our ROUGE-L/SPICE Flickr30k scores are 0.467/0.145 %(single model), 0.478/0.148 (ensemble) 
and the COCO scores are 0.549/0.194.% (single model), 0.560/0.200 (ensemble).
\label{test_result}
}}
\end{table*}
\begin{table*}[t]\footnotesize
\center
  \begin{tabular}{l c c c c c c c c c c c c c c }
   \toprule
  \multicolumn{1}{c}{} & \multicolumn{2}{c}{B-1}  & \multicolumn{2}{c}{B-2} & \multicolumn{2}{c}{B-3}  & \multicolumn{2}{c}{B-4}  & \multicolumn{2}{c}{METEOR}  & \multicolumn{2}{c}{ROUGE-L}  & \multicolumn{2}{c}{CIDEr} \\    
   \cmidrule(r){2-3}
   \cmidrule(r){4-5}
   \cmidrule(r){6-7}
   \cmidrule(r){8-9}
   \cmidrule(r){10-11}
   \cmidrule(r){12-13}
   \cmidrule(r){14-15}
	Method & c5 & c40 & c5 & c40 & c5 & c40 & c5 & c40 & c5 & c40 & c5 & c40 & c5 & c40 \\
    \midrule
    Google NIC \cite{vinyals2015show} & 0.713 & 0.895 & 0.542 & 0.802 & 0.407 & 0.694 & 0.309 & 0.587 & 0.254 & 0.346 & 0.530 & 0.682 & 0.943 & 0.946\\ 
    MS Captivator \cite{fang2015captions} & 0.715 & 0.907 & 0.543 & 0.819 & 0.407 & 0.710 & 0.308 & 0.601 & 0.248 & 0.339 & 0.526 & 0.680 & 0.931 & 0.937\\
    m-RNN \cite{mao2014deep} & 0.716 & 0.890 & 0.545 & 0.798 & 0.404 & 0.687 & 0.299 & 0.575 & 0.242 & 0.325 & 0.521 & 0.666 & 0.917 & 0.935 \\ 
    LRCN \cite{donahue2015long} & 0.718 & 0.895 & 0.548 & 0.804 & 0.409 & 0.695 & 0.306 & 0.585 & 0.247 & 0.335 & 0.528 & 0.678 & 0.921 & 0.934 \\
    Hard-Attention \cite{Xu2015show} & 0.705 & 0.881 & 0.528 & 0.779 & 0.383 & 0.658 & 0.277 & 0.537 & 0.241 & 0.322 & 0.516 & 0.654 & 0.865 & 0.893\\
    ATT-FCN \cite{you2016image} & 0.731 & 0.900 & 0.565 & 0.815 & 0.424 & 0.709 & 0.316 & 0.599 & 0.250 & 0.335 & 0.535 & 0.682 & 0.943 & 0.958\\
    ERD \cite{yang2016encode} & 0.720 & 0.900 & 0.550 & 0.812 & 0.414 & 0.705 & 0.313 & 0.597 & 0.256 & 0.347 & 0.533 & 0.686 & 0.965 & 0.969 \\
    MSM \cite{yao2016boosting} & 0.739 & 0.919 & 0.575 & 0.842 & 0.436 & 0.740 & 0.330 & 0.632 & 0.256 & 0.350 & 0.542 & 0.700 & 0.984 & 1.003 \\
	Ours-Adaptive & \textbf{0.748} & \textbf{0.920} & \textbf{0.584} & \textbf{0.845} & \textbf{0.444} & \textbf{0.744} & \textbf{0.336} & \textbf{0.637} & \textbf{0.264}&\textbf{0.359}&\textbf{0.550}&\textbf{0.705}&\textbf{1.042}&\textbf{1.059}\\
    \bottomrule
  \end{tabular}
\small{\caption{Leaderboard of the published state-of-the-art image captioning models on the online COCO testing server. Our submission is a ensemble of 5 models trained with different initialization. \label{challenge_result}
}}
\vspace{-3mm}
\end{table*}
Recently, attention mechanisms have been introduced to encoder-decoder neural frameworks in image captioning. Xu \etal \cite{Xu2015show} incorporate an attention mechanism to learn a latent alignment from scratch when generating corresponding words. \cite{wu2015value,you2016image} utilize high-level concepts or attributes and inject them into a neural-based approach as semantic attention to enhance image captioning. Yang \etal \cite{yang2016encode} extend current attention encoder-decoder frameworks using a review network, which captures the global properties in a compact vector representation and are usable by the attention mechanism in the decoder. 
Yao \etal \cite{yao2016boosting} present variants of architectures for augmenting high-level attributes from images to complement image representation for sentence generation.

To the best of our knowledge, ours is the first work to reason about when a model should attend to an image when generating a sequence of words.

%however???

%%%%%%%%%%%%%%%%%%%%%%%%%%%%%%%%%%%%%%%%%%%%%%%%%%%%%%%%%%%
\section{Results}
\label{sec:experiments}
%%%%%%%%%%%%%%%%%%%%%%%%%%%%%%%%%%%%%%%%%%%%%%%%%%%%%%%%%%%
\subsection{Experiment Settings}
We experiment with two datasets: Flickr30k \cite{young2014image} and COCO \cite{lin2014microsoft}.

\textbf{Flickr30k} contains 31,783 images collected from Flickr. Most of these images depict humans performing various activities. Each image is paired with 5 crowd-sourced captions. We use the publicly available splits\footnote{\label{note1}https://github.com/karpathy/neuraltalk} containing 1,000 images for validation and test each.

\textbf{COCO} is the largest image captioning dataset, containing 82,783, 40,504 and 40,775 images for training, validation and test respectively. This dataset is more challenging, since most images contain multiple objects in the context of complex scenes. Each image has 5 human annotated captions. For offline evaluation, we use the same data split as in \cite{karpathy2015deep,Xu2015show,you2016image} containing 5000 images for validation and test each. For online evaluation on the COCO evaluation server, we reserve 2000 images from validation for development and the rest for training. 

\textbf{Pre-processing.} We truncate captions longer than 18 words for COCO and 22 for Flickr30k. We then build a vocabulary of words that occur at least 5 and 3 times in the training set, resulting in 9567 and 7649 words for COCO and Flickr30k respectively.

\textbf{Compared Approaches:} For offline evaluation on Flickr30k and COCO, we first compare our full model (\textbf{Ours-Adaptive}) with an ablated version (\textbf{Ours-Spatial}), which only performs the spatial attention. The goal of this comparison is to verify that our improvements are not the result of orthogonal contributions (e.g. better CNN features or better optimization). We further compare our method with \textbf{DeepVS} \cite{karpathy2015deep}, \textbf{Hard-Attention} \cite{Xu2015show} and recently proposed \textbf{ATT} \cite{you2016image}, \textbf{ERD} \cite{yang2016encode} and best performed method (LSTM-A$_5$) of \textbf{MSM} \cite{yao2016boosting}. For online evaluation, we compare our method with \textbf{Google NIC} \cite{vinyals2015show}, \textbf{MS Captivator} \cite{fang2015captions}, \textbf{m-RNN} \cite{mao2014deep}, \textbf{LRCN} \cite{donahue2015long}, \textbf{Hard-Attention} \cite{Xu2015show}, \textbf{ATT-FCN} \cite{you2016image}, \textbf{ERD} \cite{yang2016encode} and \textbf{MSM} \cite{yao2016boosting}.  
\begin{figure*}[tb]
\center
\begin{minipage}{0.16\linewidth}
\includegraphics[width=1\linewidth,  height=1\linewidth, scale=0.8, trim={0 0in 0 0in}, clip]{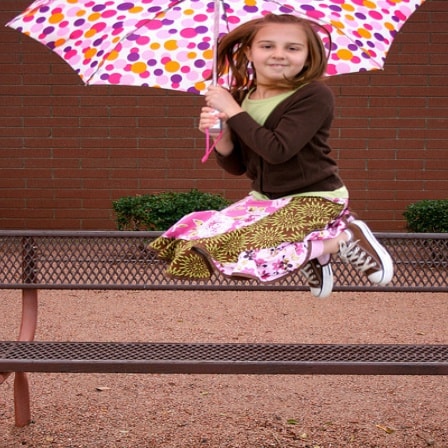}
\end{minipage}
\hspace{-1.5mm}
\begin{minipage}{0.16\linewidth}
\includegraphics[width=1\linewidth,  height=1\linewidth, scale=0.8, trim={0 0in 0 0in}, clip]{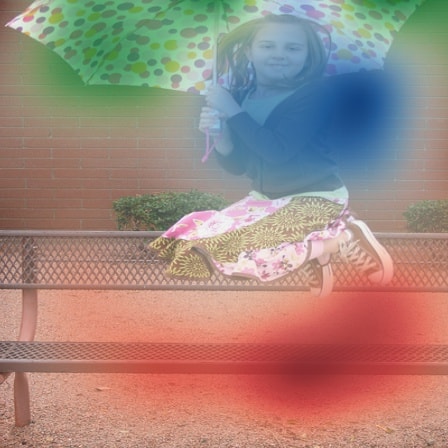}
\end{minipage}
\hspace{0.5mm}
\begin{minipage}{0.16\linewidth}
\includegraphics[width=1\linewidth,  height=1\linewidth, scale=0.8, trim={0 0in 0 0in}, clip]{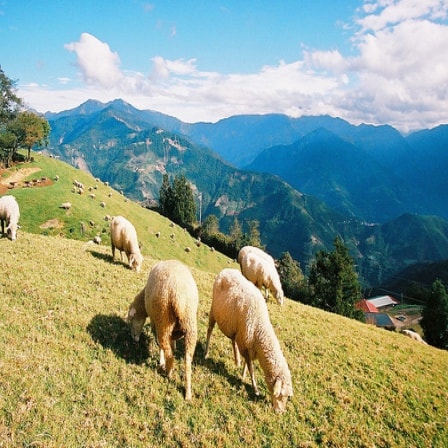}
\end{minipage}
\hspace{-1.5mm}
\begin{minipage}{0.16\linewidth}
\includegraphics[width=1\linewidth,  height=1\linewidth, scale=0.8, trim={0 0in 0 0in}, clip]{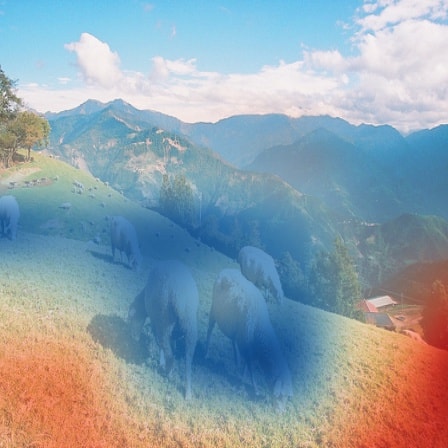}
\end{minipage}
\hspace{0.5mm}
\begin{minipage}{0.16\linewidth}
\includegraphics[width=1\linewidth,  height=1\linewidth, scale=0.8, trim={0 0in 0 0in}, clip]{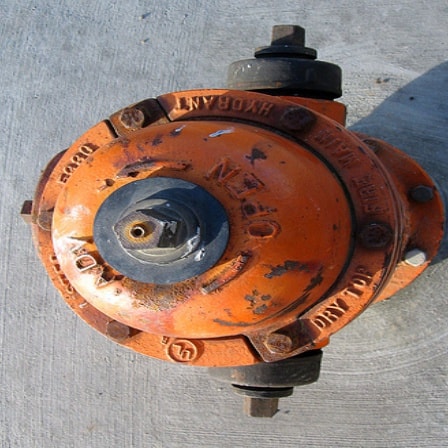}
\end{minipage}
\hspace{-1.5mm}
\begin{minipage}{0.16\linewidth}
\includegraphics[width=1\linewidth,  height=1\linewidth, scale=0.8, trim={0 0in 0 0in}, clip]{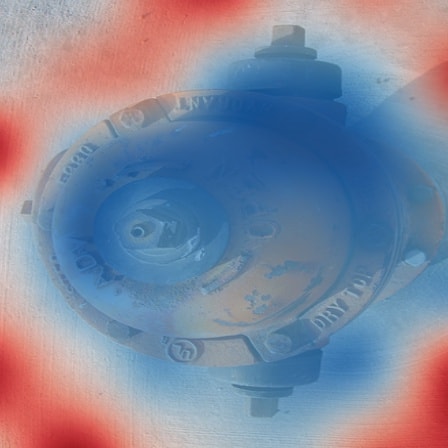}
\end{minipage}
\begin{minipage}{0.32\linewidth}
\begin{center}
\small{a little \setulcolor{blue}\ul{girl} sitting on a \setulcolor{red}\ul{bench} holding an \setulcolor{green}\ul{umbrella}.}
\end{center}
\end{minipage}
\hspace{0.5mm}
\begin{minipage}{0.32\linewidth}
\begin{center}
\small{a herd of \setulcolor{blue}\ul{sheep} grazing on a lush green \setulcolor{red}\ul{hillside}.}
\end{center}
\end{minipage}
\hspace{0.5mm}
\begin{minipage}{0.32\linewidth}
\begin{center}
\small{a close up of a  \setulcolor{blue}\ul{fire} hydrant on a \setulcolor{red}\ul{sidewalk}.}
\end{center}
\end{minipage}
\\
\vspace{1mm}
\begin{minipage}{0.16\linewidth}
\includegraphics[width=1\linewidth,  height=1\linewidth, scale=0.8, trim={0 0in 0 0in}, clip]{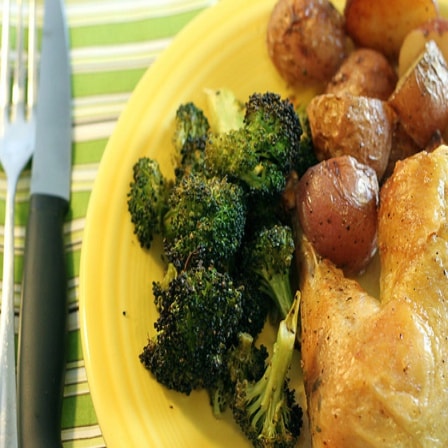}
\end{minipage}
\hspace{-1.5mm}
\begin{minipage}{0.16\linewidth}
\includegraphics[width=1\linewidth,  height=1\linewidth, scale=0.8, trim={0 0in 0 0in}, clip]{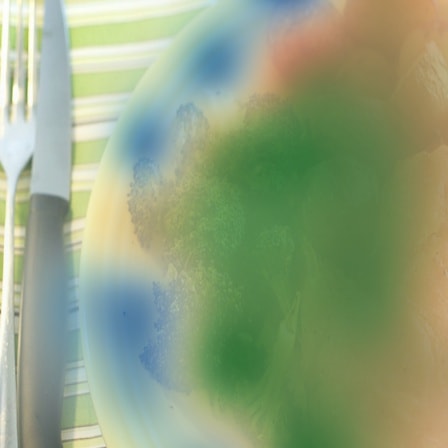}
\end{minipage}
\hspace{0.5mm}
\begin{minipage}{0.16\linewidth}
\includegraphics[width=1\linewidth,  height=1\linewidth, scale=0.8, trim={0 0in 0 0in}, clip]{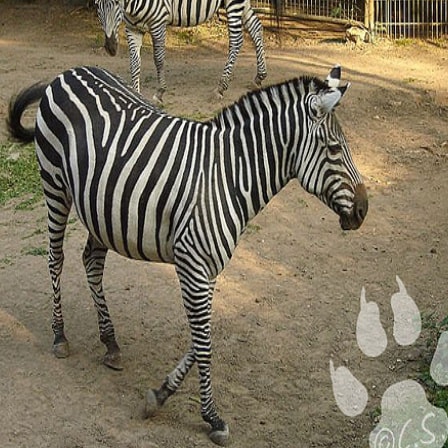}
\end{minipage}
\hspace{-1.5mm}
\begin{minipage}{0.16\linewidth}
\includegraphics[width=1\linewidth,  height=1\linewidth, scale=0.8, trim={0 0in 0 0in}, clip]{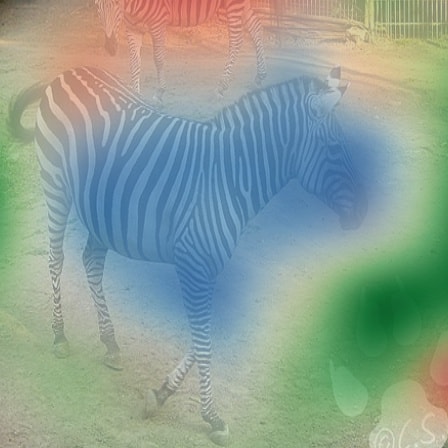}
\end{minipage}
\hspace{0.5mm}
\begin{minipage}{0.16\linewidth}
\includegraphics[width=1\linewidth,  height=1\linewidth, scale=0.8, trim={0 0in 0 0in}, clip]{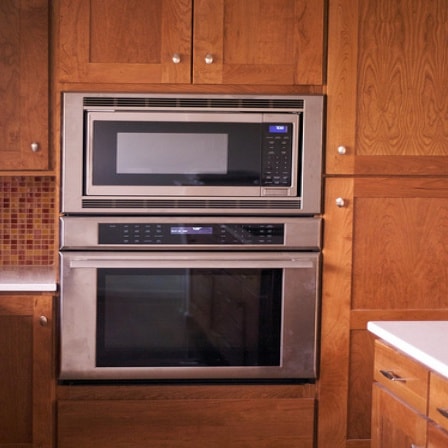}
\end{minipage}
\hspace{-1.5mm}
\begin{minipage}{0.16\linewidth}
\includegraphics[width=1\linewidth,  height=1\linewidth, scale=0.8, trim={0 0in 0 0in}, clip]{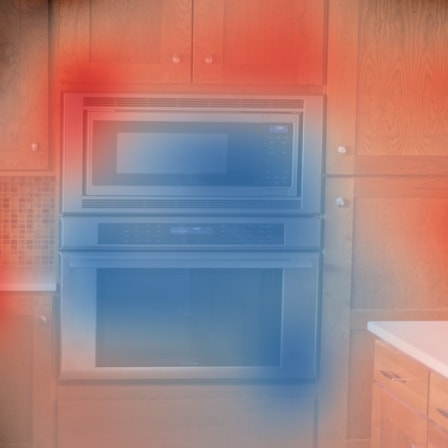}
\end{minipage}
\begin{minipage}{0.32\linewidth}
\begin{center}
\small{a yellow \setulcolor{blue}\ul{plate} topped with \setulcolor{red}\ul{meat} and \setulcolor{green}\ul{broccoli}.}
\end{center}
\end{minipage}
\hspace{0.5mm}
\begin{minipage}{0.32\linewidth}
\begin{center}
\small{a \setulcolor{blue}\ul{zebra} standing next to a \setulcolor{red}\ul{zebra} in a \setulcolor{green}\ul{dirt} field.}
\end{center}
\end{minipage}
\hspace{0.5mm}
\begin{minipage}{0.32\linewidth}
\begin{center}
\small{a \setulcolor{blue}\ul{stainless} steel oven in a kitchen with \setulcolor{red}\ul{wood} cabinets.}
\end{center}
\end{minipage}
\\
\vspace{1mm}
\begin{minipage}{0.16\linewidth}
\includegraphics[width=1\linewidth,  height=1\linewidth, scale=0.8, trim={0 0in 0 0in}, clip]{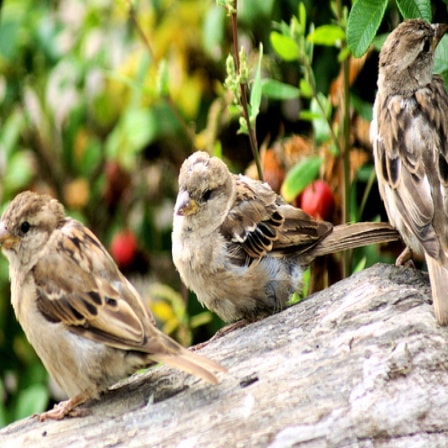}
\end{minipage}
\hspace{-1.5mm}
\begin{minipage}{0.16\linewidth}
\includegraphics[width=1\linewidth,  height=1\linewidth, scale=0.8, trim={0 0in 0 0in}, clip]{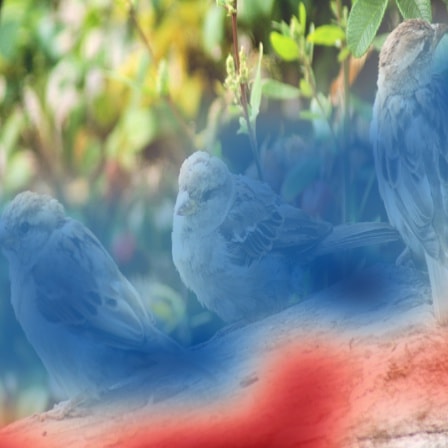}
\end{minipage}
\hspace{0.5mm}
\begin{minipage}{0.16\linewidth}
\includegraphics[width=1\linewidth,  height=1\linewidth, scale=0.8, trim={0 0in 0 0in}, clip]{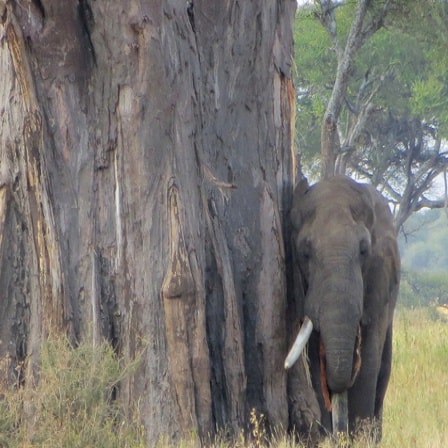}
\end{minipage}
\hspace{-1.5mm}
\begin{minipage}{0.16\linewidth}
\includegraphics[width=1\linewidth,  height=1\linewidth, scale=0.8, trim={0 0in 0 0in}, clip]{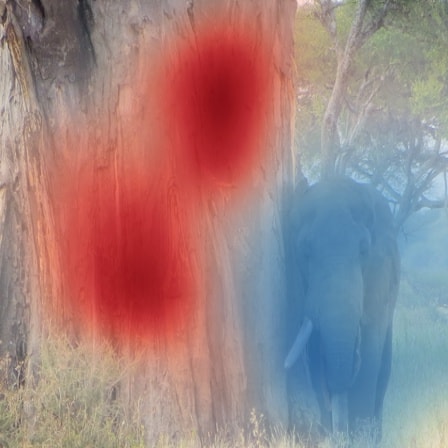}
\end{minipage}
\hspace{0.5mm}
\begin{minipage}{0.16\linewidth}
\includegraphics[width=1\linewidth,  height=1\linewidth, scale=0.8, trim={0 0in 0 0in}, clip]{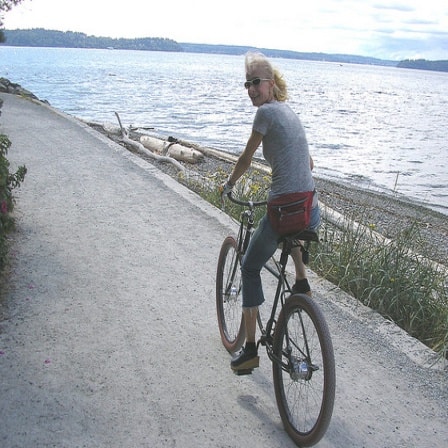}
\end{minipage}
\hspace{-1.5mm}
\begin{minipage}{0.16\linewidth}
\includegraphics[width=1\linewidth,  height=1\linewidth, scale=0.8, trim={0 0in 0 0in}, clip]{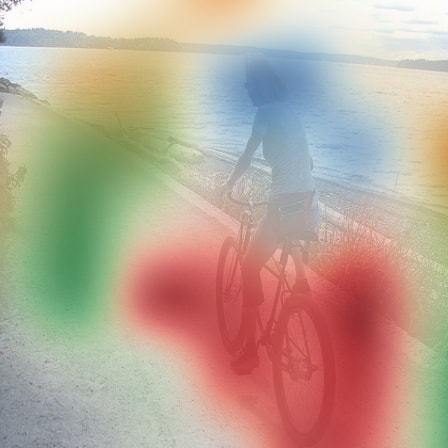}
\end{minipage}
\begin{minipage}{0.32\linewidth}
\begin{center}
\small{two \setulcolor{blue}\ul{birds} sitting on top of a \setulcolor{red}\ul{tree} branch.}
\end{center}
\end{minipage}
\hspace{0.5mm}
\begin{minipage}{0.32\linewidth}
\begin{center}
\small{an \setulcolor{blue}\ul{elephant} standing next to  rock \setulcolor{red}\ul{wall}.}
\end{center}
\end{minipage}
\hspace{0.5mm}
\begin{minipage}{0.32\linewidth}
\begin{center}
\small{a \setulcolor{blue}\ul{man} riding a \setulcolor{red}\ul{bike} down a \setulcolor{green}\ul{road} next to a body of \setulcolor{orange}\ul{water}.}
\end{center}
\end{minipage}
\captionof{figure}{Visualization of generated captions and image attention maps on the COCO dataset. Different colors show a correspondence between attended regions and underlined words. First 2 columns are success cases, last columns are failure examples. Best viewed in color. 
\label{Fig:spatialVis}
}
\vspace{-3mm}
\end{figure*}
\subsection{Quantitative Analysis} %: Caption evaluation
We report results using the COCO captioning evaluation tool \cite{lin2014microsoft}, which reports the following metrics: BLEU \cite{papineni2002bleu}, Meteor \cite{denkowski2014meteor}, Rouge-L \cite{lin2004rouge} and CIDEr \cite{vedantam2015cider}. We also report results using the new metric SPICE \cite{anderson2016spice}, which was found to better correlate with human judgments. 

Table~\ref{test_result} shows results on the Flickr30k and COCO datasets. Comparing the full model w.r.t ablated versions without visual sentinel verifies the effectiveness of the proposed framework. Our adaptive attention model significantly outperforms spatial attention model, which improves the CIDEr score from 0.493/1.029 to 0.531/1.085 on Flickr30k and COCO respectively. When comparing with previous methods, we can see that our single model significantly outperforms all previous methods in all metrics. 
On COCO, our approach improves the state-of-the-art on BLEU-4 from 0.325 (MSM$^\dagger$) to 0.332, METEOR from 0.251 (MSM$^\dagger$) to 0.266, and CIDEr from 0.986 (MSM$^\dagger$) to 1.085.
Similarly, on Flickr30k, our model improves the state-of-the-art with a large margin.

We compare our model to state-of-the-art systems on the COCO evaluation server in Table~\ref{challenge_result}. We can see that our approach achieves the best performance on all metrics among the published systems. Notably, Google NIC, ERD and MSM use Inception-v3 \cite{szegedy2015rethinking} as the encoder, which has similar or better classification performance compared to ResNet-152 \cite{he2015deep} (which is what our model uses). 

\subsection{Qualitative Analysis}
\begin{figure*}[t]
\begin{minipage}{\textwidth}
  	\begin{minipage}{0.49\textwidth} \small
   	\centering
    \includegraphics[width=1\linewidth, scale=1, trim={0 0in 0 0in}, clip]{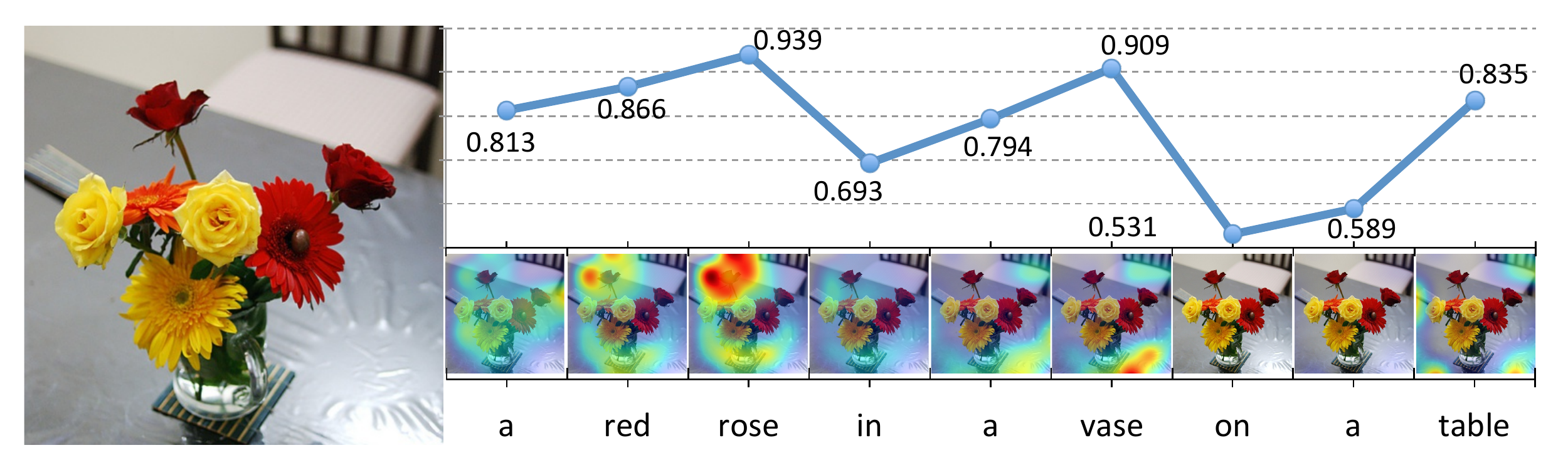} 
  \end{minipage}
  \hfill
  \begin{minipage}{0.49\textwidth}
    \centering
    \includegraphics[width=1\linewidth, scale=1, trim={0 0in 0 3 0in}, clip]{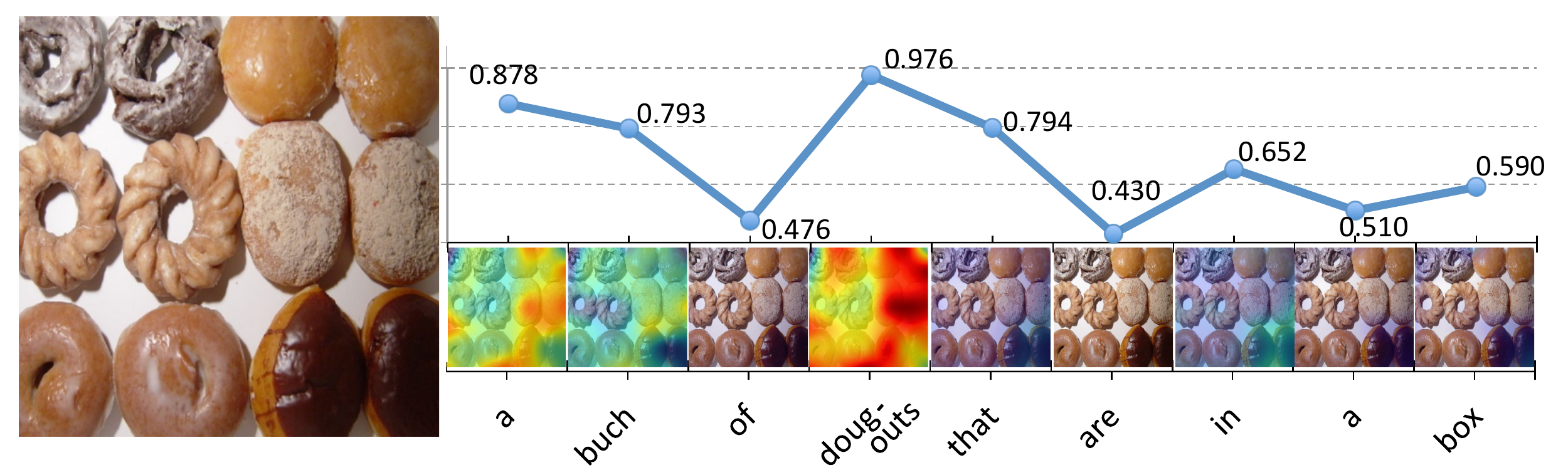}
  \end{minipage}    
\end{minipage}
\\
\begin{minipage}{\textwidth}
  	\begin{minipage}{0.49\textwidth} \small
   	\centering
    \includegraphics[width=1\linewidth, scale=1, trim={0 0in 0 0in}, clip]{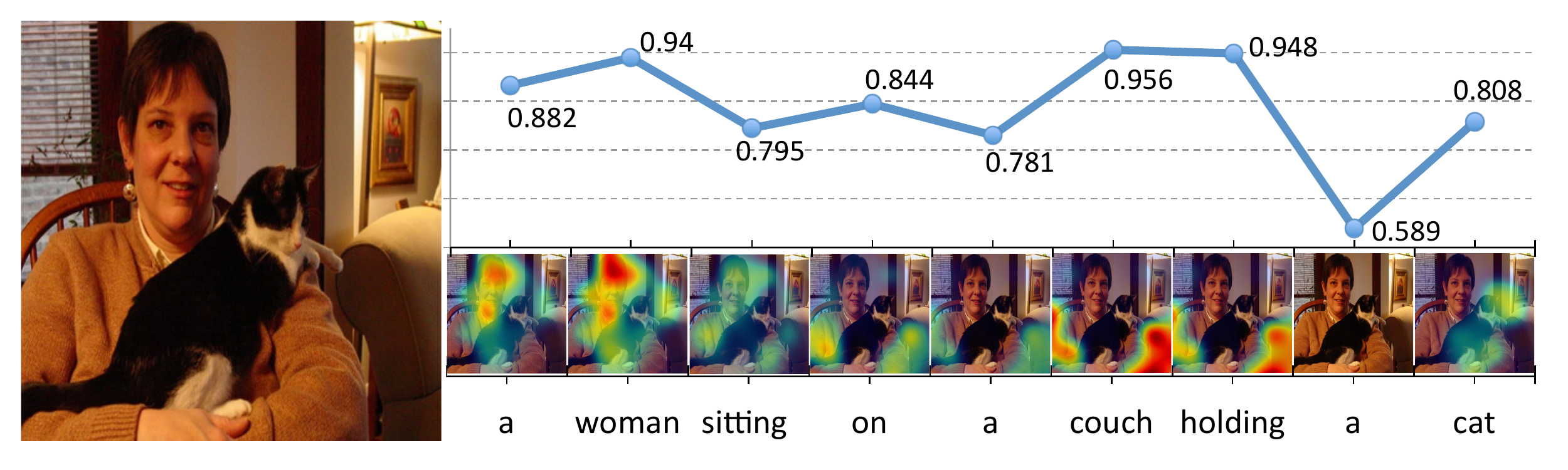} 
  \end{minipage}
  \hfill
  \begin{minipage}{0.49\textwidth}
    \centering
    \includegraphics[width=1\linewidth, scale=1, trim={0 0in 0 3 0in}, clip]{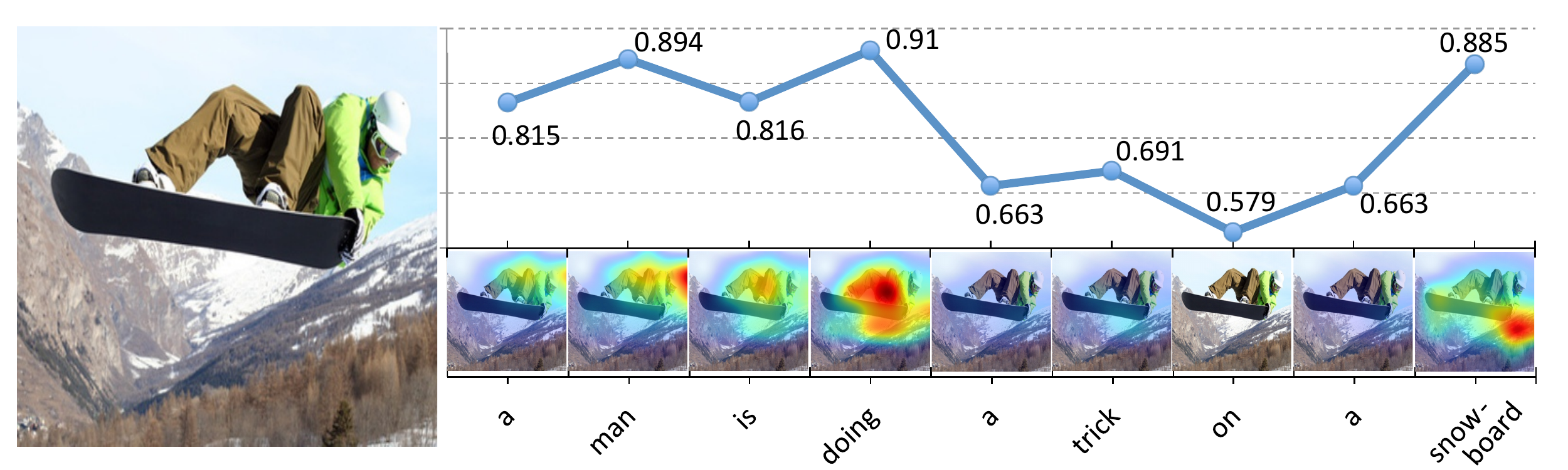}
  \end{minipage}    
\end{minipage}
\captionof{figure}{Visualization of generated captions, visual grounding probabilities of each generated word, and corresponding spatial attention maps produced by our model.}
\label{Fig:temporal}
\vspace{-3mm}
\end{figure*}

\begin{figure*}[t]
\centering
\begin{minipage}{\textwidth}
  	\begin{minipage}{0.49\textwidth} \small
   	\centering
    \includegraphics[width=1\linewidth, scale=1, trim={0 0in 0 0in}, clip]{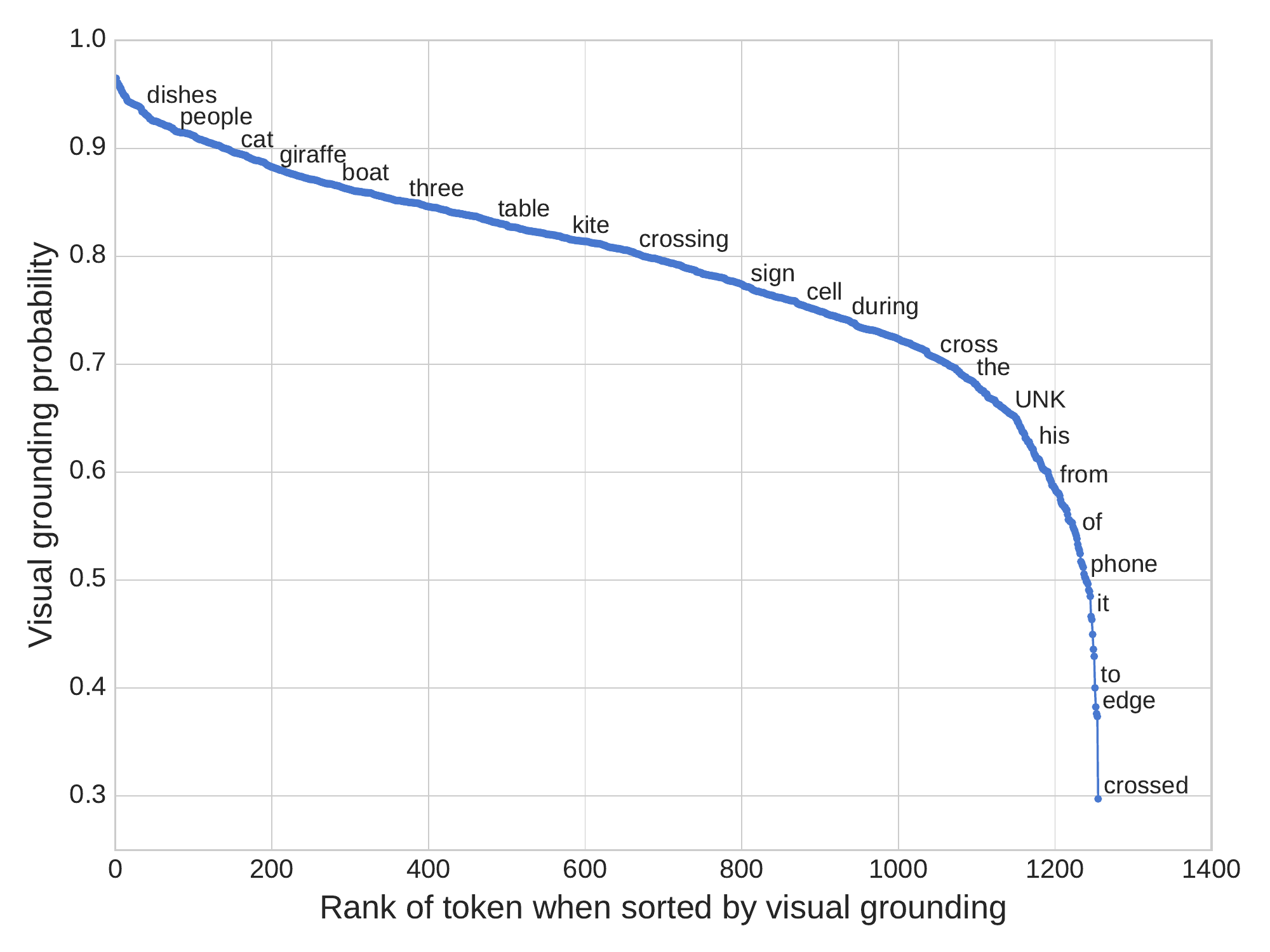} 
  \end{minipage}
  \hfill
  \begin{minipage}{0.49\textwidth}
    \centering
    \includegraphics[width=1\linewidth, scale=1, trim={0 0in 0 3 0in}, clip]{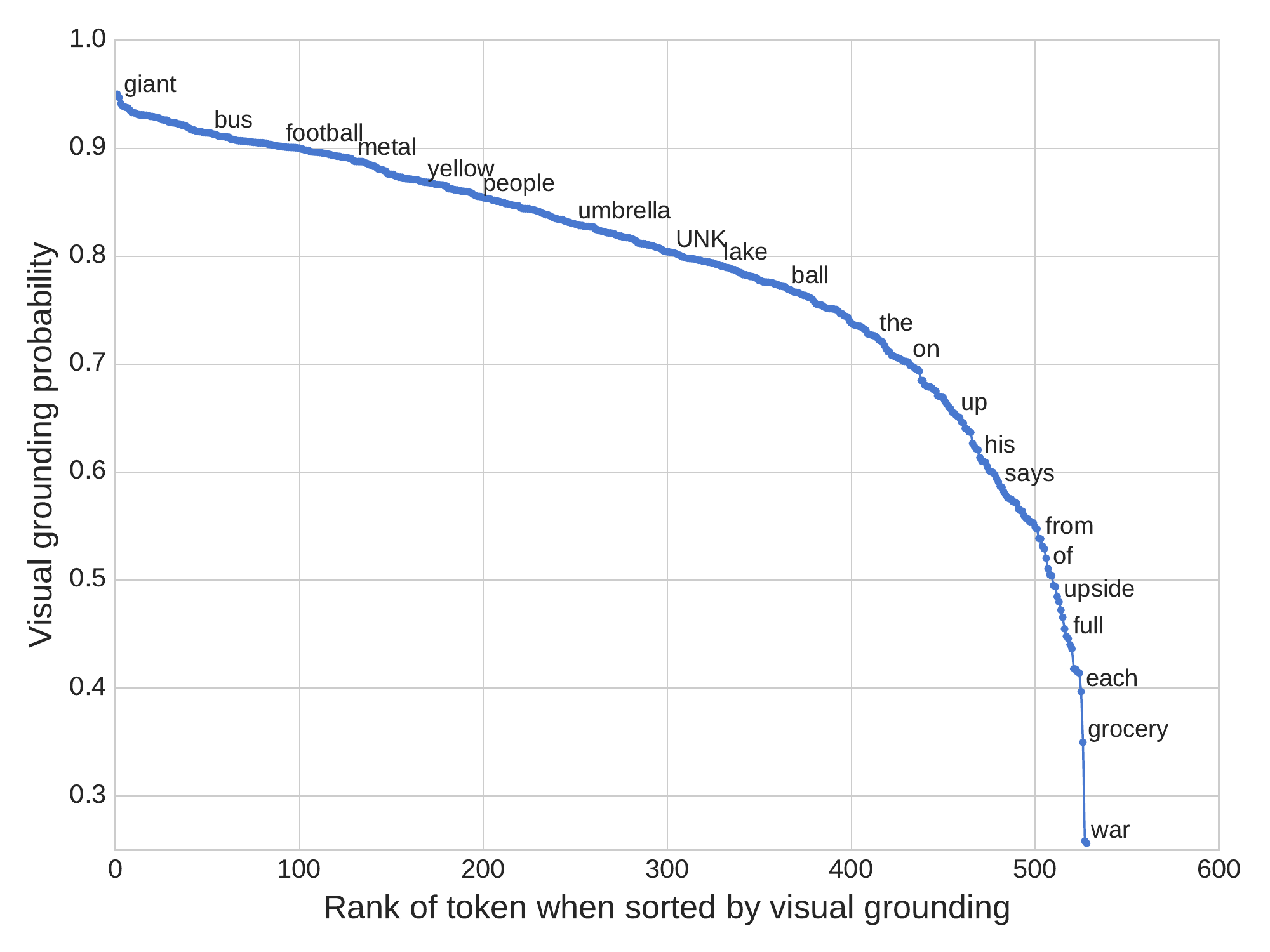}
  \end{minipage}    
\end{minipage}
\small{\caption{Rank-probability plots on COCO (left) and Flickr30k (right) indicating how likely a word is to be visually grounded when it is generated in a caption.
\label{Fig:zipf}
}}
\vspace{-3mm}
\end{figure*}
To better understand our model, we first visualize the spatial attention weight $\bm{\alpha}$ for different words in the generated caption. We simply upsample the attention weight to the image size ($224 \times 224$) using bilinear interpolation.
Fig.~\ref{Fig:spatialVis} shows generated captions and the spatial attention maps for specific words in the caption. First two columns are success examples and the last one column shows failure examples. We see that our model learns alignments that correspond strongly with human intuition. 
Note that even in cases where the model produces inaccurate captions, we see that our model does look at reasonable regions in the image -- it just seems to not be able to count or recognize texture and fine-grained categories. We provide a more extensive list of visualizations in supplementary material.

We further visualize the sentinel gate as a caption is generated. For each word, we use $1-\beta$ as its visual grounding probability. In Fig.~\ref{Fig:temporal}, we visualize the generated caption, the visual grounding probability and the spatial attention map generated by our model for each word. Our model successfully learns to attend to the image less when generating non-visual words such as ``of'' and ``a''. For visual words like ``red'', ``rose'', ``doughnuts'', ``woman'' and ``snowboard'', our model assigns a high visual grounding probabilities (over 0.9). Note that the same word may be assigned different visual grounding probabilities when generated in different contexts. For example, the word ``a'' usually has a high visual grounding probability at the beginning of a sentence, since without any language context, the model needs the visual information to determine plurality (or not). On the other hand, the visual grounding probability of "a" in the phrase ``on a table'' is much lower. Since it is unlikely for something to be on more than one table.
\subsection{Adaptive Attention Analysis}
\label{exp:aaa}
\begin{figure*}[t]\footnotesize
\center
\includegraphics[width=0.9\linewidth, scale=1, trim={0 0in 0 0in}, clip]{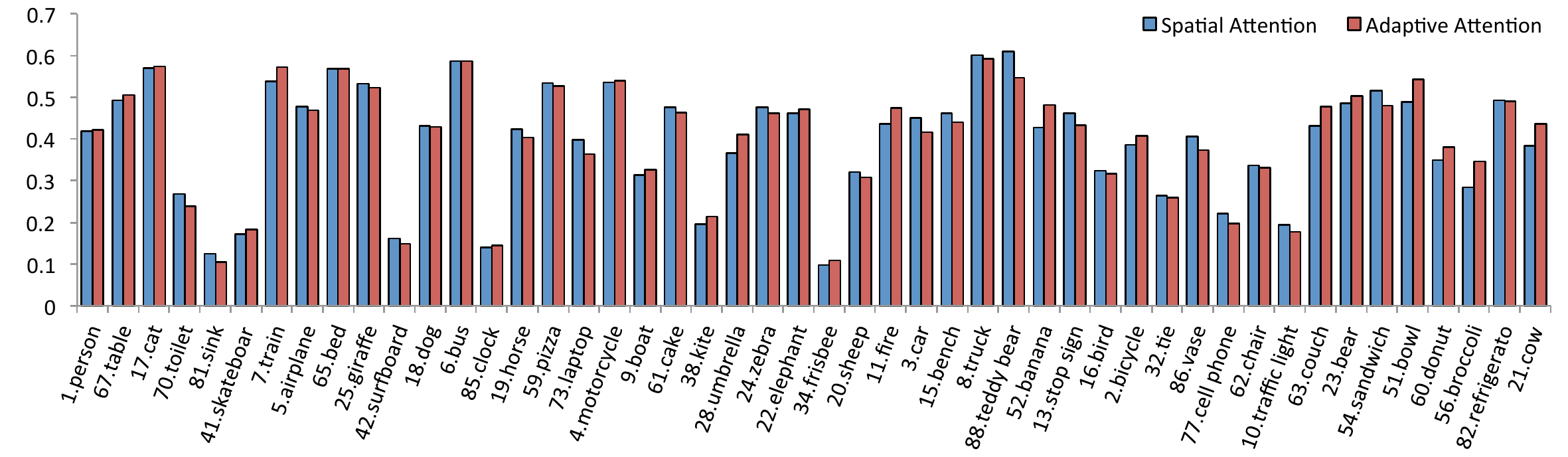} 
\small{\caption{Localization accuracy over generated captions for top 45 most frequent COCO object categories. ``Spatial Attention'' and ``Adaptive Attention'' are our proposed spatial attention model and adaptive attention model, respectively. The COCO categories are ranked based on the align results of our adaptive attention, which cover 93.8\% and 94.0\% of total matched regions for spatial attention and adaptive attention, respectively.
\label{fig:weakly_local}
}}
\vspace{-3mm}
\end{figure*}
%overview, say our evaluation is on the generated captions., how to evaluate them. 
In this section, we analysis the adaptive attention generated by our methods. We visualize the sentinel gate to understand ``when'' our model attends to the image as a caption is generated. We also perform a weakly-supervised localization 
on COCO categories by using the generated attention maps. This can help us to get an intuition of ``where'' our model attends, and whether it attends to the correct regions. 
\subsubsection{Learning ``when'' to attend}
\label{sec:when}
\vspace{-1mm}
In order to assess whether our model learns to separate visual words in captions from non-visual words, we visualize the visual grounding probability. For each word in the vocabulary, we average the visual grounding probability over all the generated captions containing that word. Fig.~\ref{Fig:zipf} shows the rank-probability plot on COCO and Flickr30k. 

We find that our model attends to the image more when generating object words like ``dishes'', ``people'', ``cat'', ``boat''; attribute words like ``giant'', ``metal'', ``yellow'' and number words like ``three''. When the word is non-visual, our model learns to not attend to the image such as for ``the'', ``of'', ``to'' etc. For more abstract notions such as ``crossing'', ``during'' etc., our model leans to attend less than the visual words and attend more than the non-visual words. Note that our model does not rely on any syntactic features or external knowledge. It discovers these trends automatically. % which is much more flexible.

Our model cannot distinguish between words that are truly non-visual from the ones that are technically visual but have a high correlation with other words and hence chooses to not rely on the visual signal. For example, words such as ``phone'' get a relatively low visual grounding probability in our model. This is because it has a large language correlation with the word ``cell''. We can also observe some interesting trends in what the model learns on different datasets. For example, when generating ``UNK'' words, our model learns to attend less to the image on COCO, but more on Flickr30k. Same words with different forms can also results in different visual grounding probabilities. For example, ``crossing'', ``cross'' and ``crossed'' are cognate words which have similar meaning. However, in terms of the visual grounding probability learnt by our model, there is a large variance. Our model learns to attend to images more when generating ``crossing'', followed by ``cross'' and attend least on image when generating ``crossed''. %\devi{Does this make sense though?}
\vspace{-1mm}
\subsubsection{Learning ``where'' to attend}
%Note that this is the first time using the caption attention on this task. Discuss the with the attention correctness. 
\label{sec:where}
\vspace{-1mm}
We now assess whether our model attends to the correct spatial image regions. We perform weakly-supervised localization \cite{ram2016grad, zhou2015learning} using the generated attention maps. To the best of our best knowledge, no previous works have used weakly supervised localization to evaluate spatial attention for image captioning. Given the word $w_t$ and attention map $\bm{\alpha}_t$, we first segment the regions of of the image with attention values larger than $th$ (after map is normalized to have the largest value be 1), where $th$ is a per-class threshold estimated using the COCO validation split. Then we take the bounding box that covers the largest connected component in the segmentation map. 
We use intersection over union (IOU) of the generated and ground truth bounding box as the localization accuracy. 

For each of the COCO object categories, we do a word-by-word match to align the generated words with the ground truth bounding box. For the object categories which has multiple words, such as ``teddy bear'', we take the maximum IOU score over the multiple words as its localization accuracy. 
%\devi{But why can't you manually just map teddy bear to some word?}
%
We are able to align 5981 and 5924 regions for captions generated by the spatial and adaptive attention models respectively. 
The average localization accuracy for our spatial attention model is \textbf{0.362}, and \textbf{0.373} for our adaptive attention model. This demonstrates that as a byproduct, knowing when to attend also helps where to attend. %By adding the adaptive attention mechanism, our model performs better on weakly supervised localization task. This shows knowing when to attend can also help knowing where to attend. 

Fig.~\ref{fig:weakly_local} shows the localization accuracy over the generated captions for top 45 most frequent COCO object categories. We can see that our spatial attention and adaptive attention models share similar trends. We observe that both models perform well on categories such as ``cat'', ``bed'', ``bus'' and ``truck''. On smaller objects, such as ``sink'', ``surfboard'', ``clock'' and ``frisbee'', both models perform relatively poorly. This is because our spatial attention maps are directly rescaled from a coarse $7\times 7$ feature map, which looses a lot of spatial resolution and detail. Using a larger feature map may improve the performance. 
%
%%%%%%%%%%%%%%%%%%%%%%%%%%%%%%%%%%%%%%%%%%%%%%%%%%%%%%%%%%%
\section{Conclusion}
\label{sec:conclusion}
\vspace{-1mm}
%%%%%%%%%%%%%%%%%%%%%%%%%%%%%%%%%%%%%%%%%%%%%%%%%%%%%%%%%%%
In this paper, we present a novel adaptive attention encoder-decoder framework, which provides a fallback option to the decoder. We further introduce a new LSTM extension, which produces an additional ``visual sentinel''. Our model achieves state-of-the-art performance across standard benchmarks on image captioning. We perform extensive attention evaluation to analysis our adaptive attention. Though our model is evaluated on image captioning, it can have useful applications in other domains.

\small{\textbf{Acknowledgements}}
\small{This work was funded in part by an NSF CAREER award, ONR
YIP award, Sloan Fellowship, ARO YIP award, Allen Distinguished
Investigrator award from the Paul G. Allen Family
Foundation, Google Faculty Research Award, Amazon
Academic Research Award to DP
}

{\small
\bibliographystyle{ieee}
\bibliography{egbib}
}
\section{Supplementary}
\subsection{COCO Categories Mapping List for Weakly-Supervised Localization}
We first use WordNetLemmatizer from NLTK\footnote{http://www.nltk.org/} to lemmatize each word of the caption. Then we map ``people'', ``woman'', ``women'', ``boy'', ``girl'', ``man'', ``men'', ``player'',``baby'' to COCO ``\textbf{person}'' category; ``plane'', ``jetliner'', ``jet'' to COCO ``\textbf{airplane}'' category; ``bike'' to COCO ``\textbf{bicycle}'' category; ``taxi'' to COCO ``\textbf{car}'' category. We also change the COCO category name from ``\textbf{dining table}'' to ``table'' while evaluation. For the rest categories, we keep their original names. We show the visualization of bounding box in Fig.~\ref{img_compare}

\subsection{Analysis on the gradient of non-visual words}
\begin{figure}[h]
\center
\begin{minipage}{0.16\linewidth}
\includegraphics[width=1\linewidth,  height=1\linewidth, scale=0.8, trim={0 0in 0 0in}, clip]{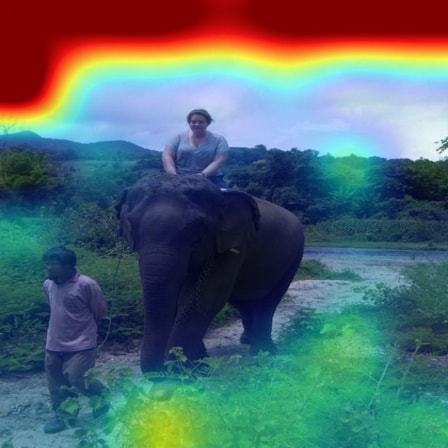}
\end{minipage}
\hspace{-1.5mm}
\begin{minipage}{0.16\linewidth}
\includegraphics[width=1\linewidth,  height=1\linewidth, scale=0.8, trim={0 0in 0 0in}, clip]{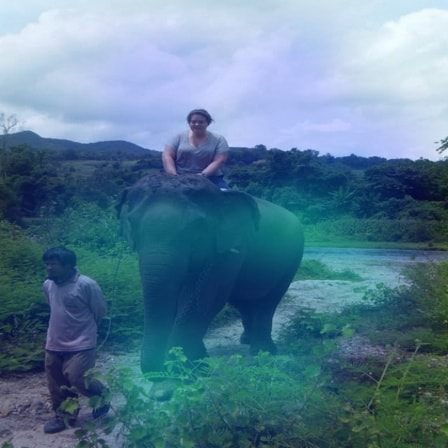}
\end{minipage}
\begin{minipage}{0.16\linewidth}
\includegraphics[width=1\linewidth,  height=1\linewidth, scale=0.8, trim={0 0in 0 0in}, clip]{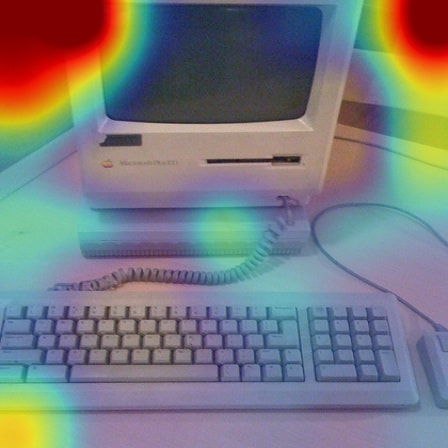}
\end{minipage}
\hspace{-1.5mm}
\begin{minipage}{0.16\linewidth}
\includegraphics[width=1\linewidth,  height=1\linewidth, scale=0.8, trim={0 0in 0 0in}, clip]{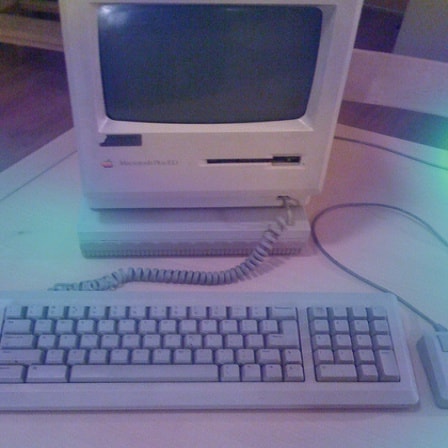}
\end{minipage}
\begin{minipage}{0.16\linewidth}
\includegraphics[width=1\linewidth,  height=1\linewidth, scale=0.8, trim={0 0in 0 0in}, clip]{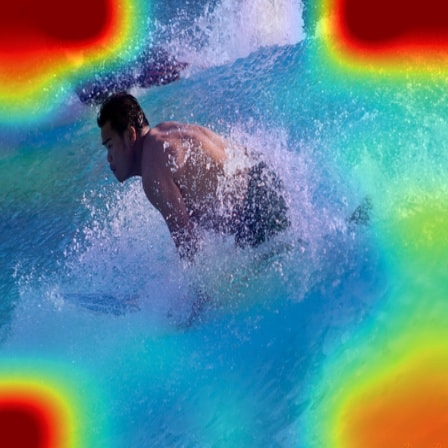}
\end{minipage}
\hspace{-1.5mm}
\begin{minipage}{0.16\linewidth}
\includegraphics[width=1\linewidth,  height=1\linewidth, scale=0.8, trim={0 0in 0 0in}, clip]{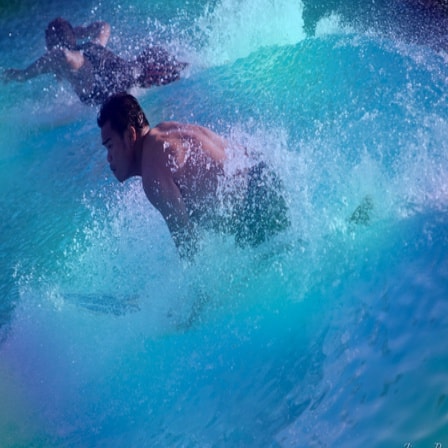}
\end{minipage}
%
%\vspace{10mm}
%
\begin{minipage}{0.16\linewidth}
\includegraphics[width=1\linewidth,  height=1\linewidth, scale=0.8, trim={0 0in 0 0in}, clip]{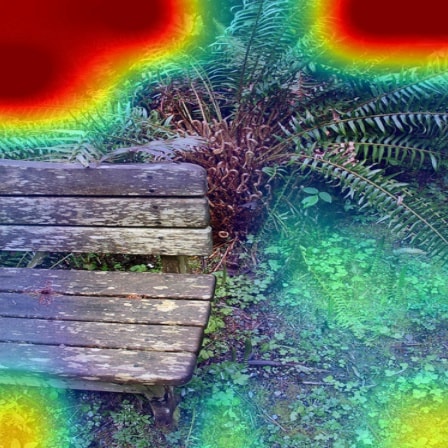}
\end{minipage}
\hspace{-1.5mm}
\begin{minipage}{0.16\linewidth}
\includegraphics[width=1\linewidth,  height=1\linewidth, scale=0.8, trim={0 0in 0 0in}, clip]{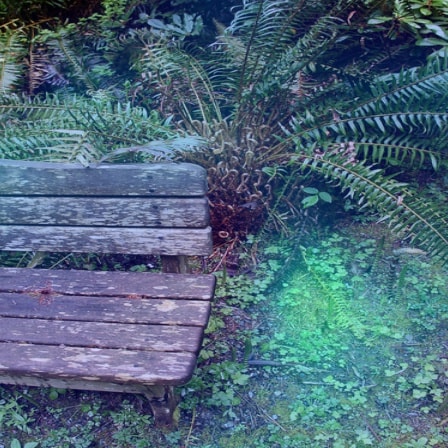}
\end{minipage}
\begin{minipage}{0.16\linewidth}
\includegraphics[width=1\linewidth,  height=1\linewidth, scale=0.8, trim={0 0in 0 0in}, clip]{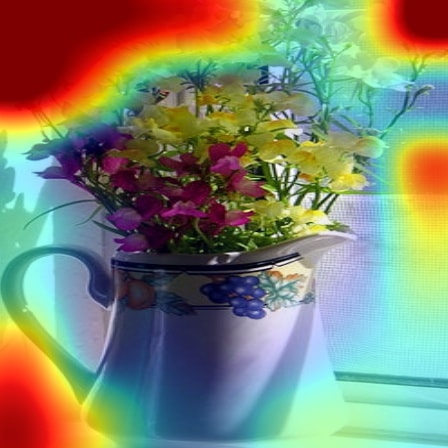}
\end{minipage}
\hspace{-1.5mm}
\begin{minipage}{0.16\linewidth}
\includegraphics[width=1\linewidth,  height=1\linewidth, scale=0.8, trim={0 0in 0 0in}, clip]{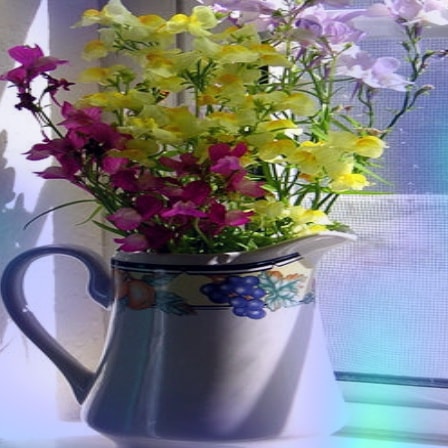}
\end{minipage}
\begin{minipage}{0.16\linewidth}
\includegraphics[width=1\linewidth,  height=1\linewidth, scale=0.8, trim={0 0in 0 0in}, clip]{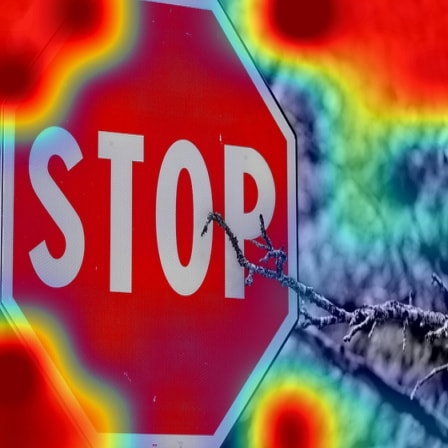}
\end{minipage}
\hspace{-1.5mm}
\begin{minipage}{0.16\linewidth}
\includegraphics[width=1\linewidth,  height=1\linewidth, scale=0.8, trim={0 0in 0 0in}, clip]{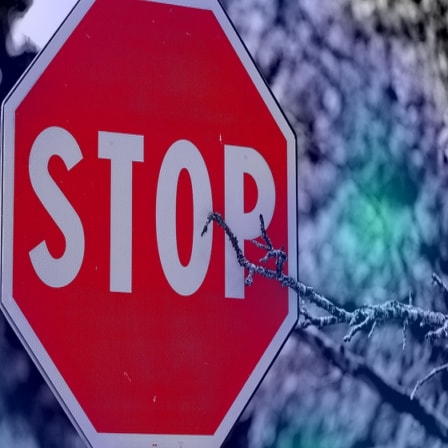}
\end{minipage}
\small{\caption{Image attention visualization of word ``of'' on several images. For each image pair, left: output of spatial attention model (no visual sentinel), right: output of our adaptive attention model (with visual sentinel). 
\label{img_compare}
}}
\end{figure}

In the experiments in Table~\ref{test_result} in the main paper, we show the effectiveness of visual sentinel in the ablation study comparing spatial attention (no visual sentinel) vs. spatial attention+visual sentinel. To further demonstrate the intuition, we have run additional experiments. In Fig.~\ref{img_compare} we see that without visual sentinel, the attention for the  non-visual word ``of'' spreads around the boundary (corner) of image. Clearly, this would result in a noisy signal being propagated through the network. Interestingly, the visual grounding probability for ``of'' in our model (with visual sentinel) is small. This restricts the noisy signal from Fig.~\ref{img_compare} from backpropagating to the visual attention model.
\subsection{Adaptive attention across different datasets}
We show the visual grounding probability for the same words across COCO and Flickr30 datasets in Table~\ref{vgp}. Trends are generally similar between the two datasets. To quantify this, we sort all common words between the two datasets by their visual grounding probabilities from both datasets. The rank correlation is 0.483. Words like ``sheep'' and ``railing'' have high visual grounding in COCO but not in Flickr30K, while ``hair'' and ``run'' are the reverse. Apart from different distributions of visual entities present in the dataset, some differences may be a consequence of different amounts of training data. Will add this to the paper.

\begin{table}[t]\footnotesize
\setlength{\tabcolsep}{1.8pt}
\center
\begin{tabular}{l c c c c c c c c c c c c c c}
    \toprule
Dataset & \rot{giant} & \rot{people} &\rot{bus} & \rot{metal} & \rot{umbrella} & \rot{lake} & \rot{yellow} &\rot{on} &\rot{the} &\rot{UNK} &\rot{full} & \rot{says} &\rot{of} &\rot{up}\\
\midrule
COCO  &\rot{0.921} & \rot{0.917} & \rot{0.868} & \rot{0.856} & \rot{0.843} & \rot{0.837} & \rot{0.827} & \rot{0.713} & \rot{0.685} & \rot{0.654} & \rot{0.622} & \rot{0.612} & \rot{0.541} & \rot{0.527}\\
   Flickr30K & \rot{0.947} & \rot{0.856} & \rot{0.914} & \rot{0.889} & \rot{0.830} & \rot{0.791} & \rot{0.869} & \rot{0.702} & \rot{0.726} & \rot{0.803} & \rot{0.445} & \rot{0.586} & \rot{0.510} & \rot{0.652}\\
    \bottomrule
\end{tabular}
\small{\caption{Visual grounding probabilities of the same word on COCO and Flickr30K datasets.
\label{vgp}
}}
\vspace{-2mm}
\end{table}

\subsection{More Visualization of Attention}

\begin{figure*}[tb]
\center
\begin{minipage}{0.16\linewidth}
\includegraphics[width=1\linewidth,  height=1\linewidth, scale=0.8, trim={0 0in 0 0in}, clip]{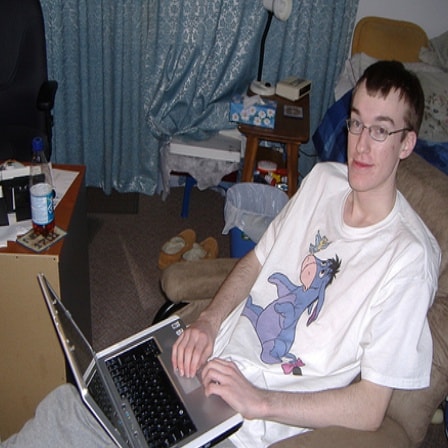}
\end{minipage}
\hspace{-1.5mm}
\begin{minipage}{0.16\linewidth}
\includegraphics[width=1\linewidth,  height=1\linewidth, scale=0.8, trim={0 0in 0 0in}, clip]{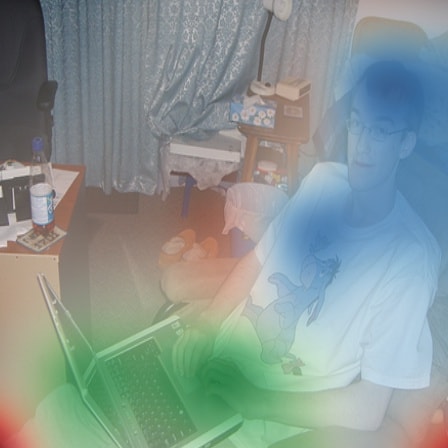}
\end{minipage}
\hspace{0.5mm}
\begin{minipage}{0.16\linewidth}
\includegraphics[width=1\linewidth,  height=1\linewidth, scale=0.8, trim={0 0in 0 0in}, clip]{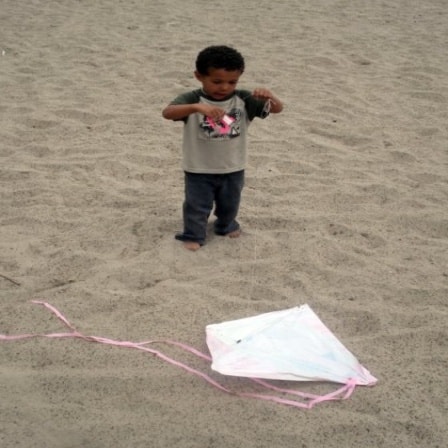}
\end{minipage}
\hspace{-1.5mm}
\begin{minipage}{0.16\linewidth}
\includegraphics[width=1\linewidth,  height=1\linewidth, scale=0.8, trim={0 0in 0 0in}, clip]{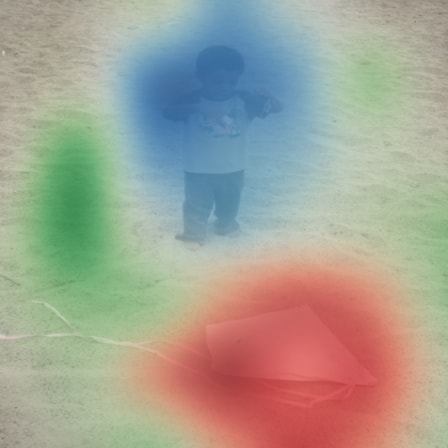}
\end{minipage}
\hspace{0.5mm}
\begin{minipage}{0.16\linewidth}
\includegraphics[width=1\linewidth,  height=1\linewidth, scale=0.8, trim={0 0in 0 0in}, clip]{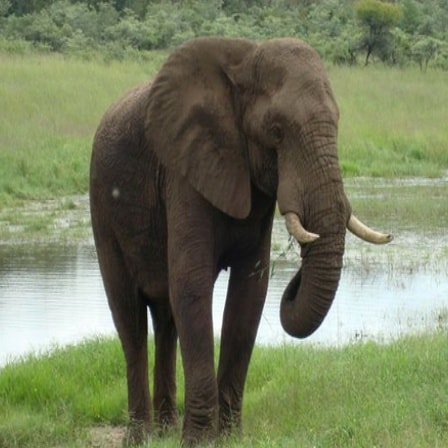}
\end{minipage}
\hspace{-1.5mm}
\begin{minipage}{0.16\linewidth}
\includegraphics[width=1\linewidth,  height=1\linewidth, scale=0.8, trim={0 0in 0 0in}, clip]{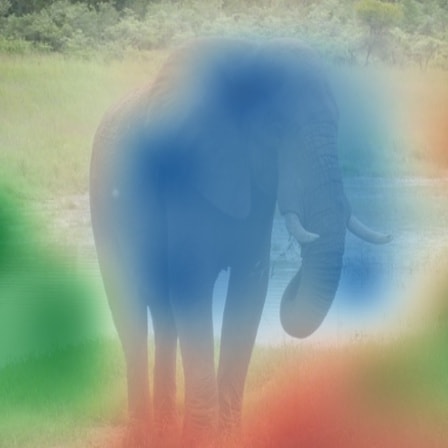}
\end{minipage}
\begin{minipage}{0.32\linewidth}
\begin{center}
\small{a man \setulcolor{blue}\ul{sitting} on a \setulcolor{red}\ul{couch} using a \setulcolor{green}\ul{laptop} computer.}
\end{center}
\end{minipage}
\hspace{0.5mm}
\begin{minipage}{0.32\linewidth}
\begin{center}
\small{a young boy \setulcolor{blue}\ul{holding} a  \setulcolor{red}\ul{kite} on a \setulcolor{green}\ul{beach}.}
\end{center}
\end{minipage}
\hspace{0.5mm}
\begin{minipage}{0.32\linewidth}
\begin{center}
\small{an \setulcolor{blue}\ul{elephant} standing in the \setulcolor{red}\ul{grass} near a \setulcolor{green}\ul{lake}.}
\end{center}
\end{minipage}
\\
\vspace{1mm}
\begin{minipage}{0.16\linewidth}
\includegraphics[width=1\linewidth,  height=1\linewidth, scale=0.8, trim={0 0in 0 0in}, clip]{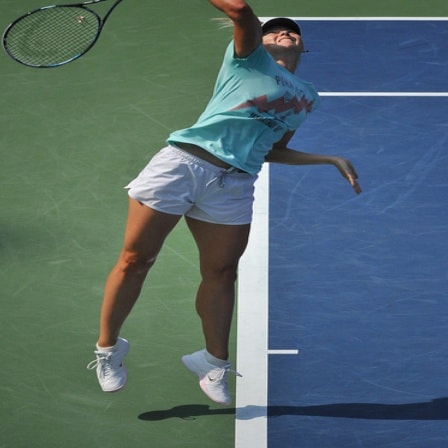}
\end{minipage}
\hspace{-1.5mm}
\begin{minipage}{0.16\linewidth}
\includegraphics[width=1\linewidth,  height=1\linewidth, scale=0.8, trim={0 0in 0 0in}, clip]{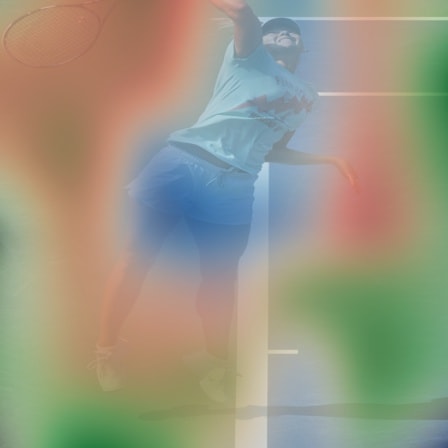}
\end{minipage}
\hspace{0.5mm}
\begin{minipage}{0.16\linewidth}
\includegraphics[width=1\linewidth,  height=1\linewidth, scale=0.8, trim={0 0in 0 0in}, clip]{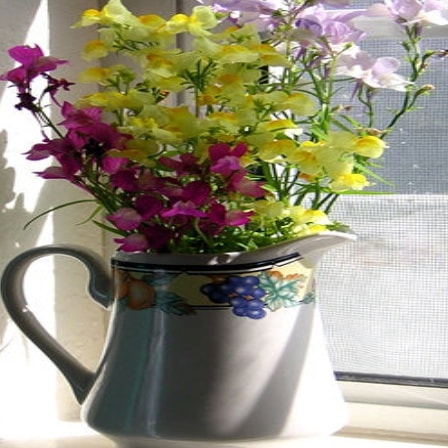}
\end{minipage}
\hspace{-1.5mm}
\begin{minipage}{0.16\linewidth}
\includegraphics[width=1\linewidth,  height=1\linewidth, scale=0.8, trim={0 0in 0 0in}, clip]{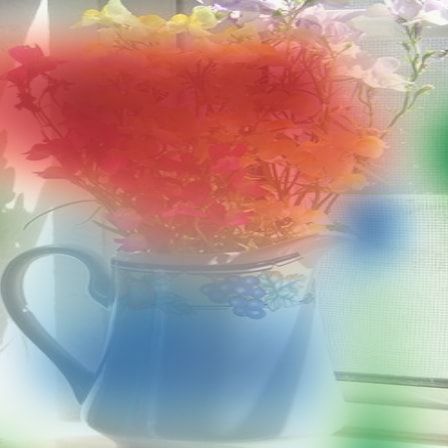}
\end{minipage}
\hspace{0.5mm}
\begin{minipage}{0.16\linewidth}
\includegraphics[width=1\linewidth,  height=1\linewidth, scale=0.8, trim={0 0in 0 0in}, clip]{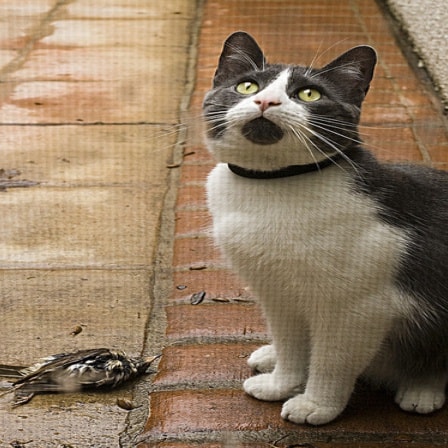}
\end{minipage}
\hspace{-1.5mm}
\begin{minipage}{0.16\linewidth}
\includegraphics[width=1\linewidth,  height=1\linewidth, scale=0.8, trim={0 0in 0 0in}, clip]{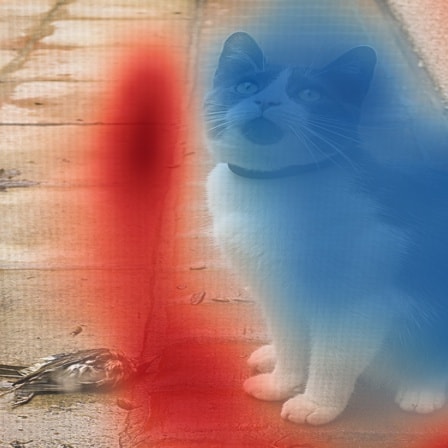}
\end{minipage}
\begin{minipage}{0.32\linewidth}
\begin{center}
\small{a \setulcolor{blue}\ul{woman} is playing \setulcolor{red}\ul{tennis} on a tennis  \setulcolor{green}\ul{court}.}
\end{center}
\end{minipage}
\hspace{0.5mm}
\begin{minipage}{0.32\linewidth}
\begin{center}
\small{a \setulcolor{blue}\ul{vase} filled with \setulcolor{red}\ul{flowers} sitting on top of a \setulcolor{green}\ul{table}.}
\end{center}
\end{minipage}
\hspace{0.5mm}
\begin{minipage}{0.32\linewidth}
\begin{center}
\small{a black and white \setulcolor{blue}\ul{cat} sitting on a brick \setulcolor{red}\ul{wall}.}
\end{center}
\end{minipage}
\\
\vspace{1mm}
\begin{minipage}{0.16\linewidth}
\includegraphics[width=1\linewidth,  height=1\linewidth, scale=0.8, trim={0 0in 0 0in}, clip]{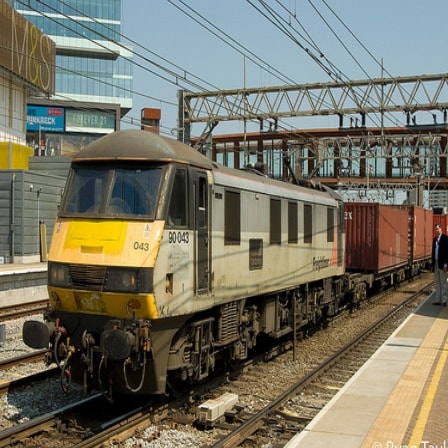}
\end{minipage}
\hspace{-1.5mm}
\begin{minipage}{0.16\linewidth}
\includegraphics[width=1\linewidth,  height=1\linewidth, scale=0.8, trim={0 0in 0 0in}, clip]{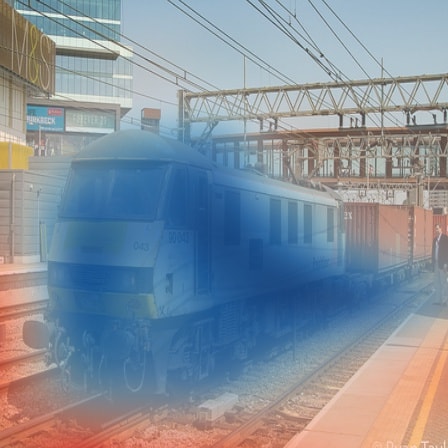}
\end{minipage}
\hspace{0.5mm}
\begin{minipage}{0.16\linewidth}
\includegraphics[width=1\linewidth,  height=1\linewidth, scale=0.8, trim={0 0in 0 0in}, clip]{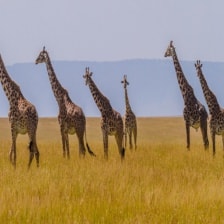}
\end{minipage}
\hspace{-1.5mm}
\begin{minipage}{0.16\linewidth}
\includegraphics[width=1\linewidth,  height=1\linewidth, scale=0.8, trim={0 0in 0 0in}, clip]{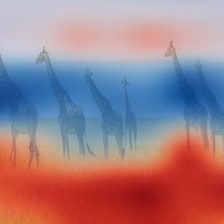}
\end{minipage}
\hspace{0.5mm}
\begin{minipage}{0.16\linewidth}
\includegraphics[width=1\linewidth,  height=1\linewidth, scale=0.8, trim={0 0in 0 0in}, clip]{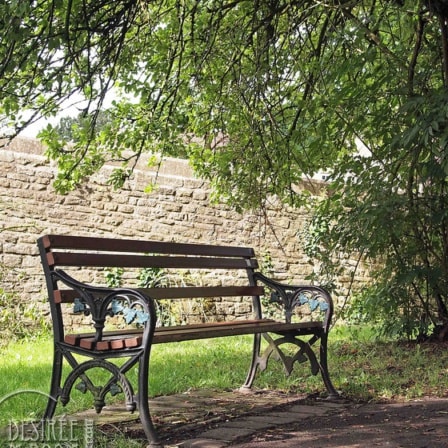}
\end{minipage}
\hspace{-1.5mm}
\begin{minipage}{0.16\linewidth}
\includegraphics[width=1\linewidth,  height=1\linewidth, scale=0.8, trim={0 0in 0 0in}, clip]{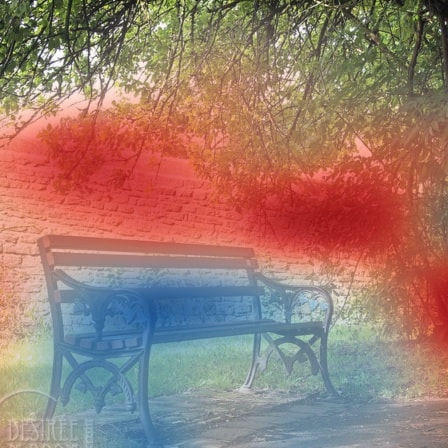}
\end{minipage}
\\
\vspace{1mm}
\begin{minipage}{0.32\linewidth}
\begin{center}
\small{a yellow and black \setulcolor{blue}\ul{train} traveling down train \setulcolor{red}\ul{track}.}
\end{center}
\end{minipage}
\hspace{0.5mm}
\begin{minipage}{0.32\linewidth}
\begin{center}
\small{as group of \setulcolor{blue}\ul{giraffes} standing in a \setulcolor{red}\ul{field}.}
\end{center}
\end{minipage}
\hspace{0.5mm}
\begin{minipage}{0.32\linewidth}
\begin{center}
\small{a wooden \setulcolor{blue}\ul{bench} sitting next to a stone \setulcolor{red}\ul{wall}.}
\end{center}
\end{minipage}
\captionof{figure}{Visualization of generated captions and image attention maps on the COCO dataset. Different colors show a correspondence between attended regions and underlined words.}
\vspace{-3mm}
\label{vis1}
\end{figure*}

\begin{figure*}[tb]
\center
\begin{minipage}{0.1\linewidth}
\includegraphics[width=1\linewidth,  height=1\linewidth, scale=0.8, trim={0 0in 0 0in}, clip]{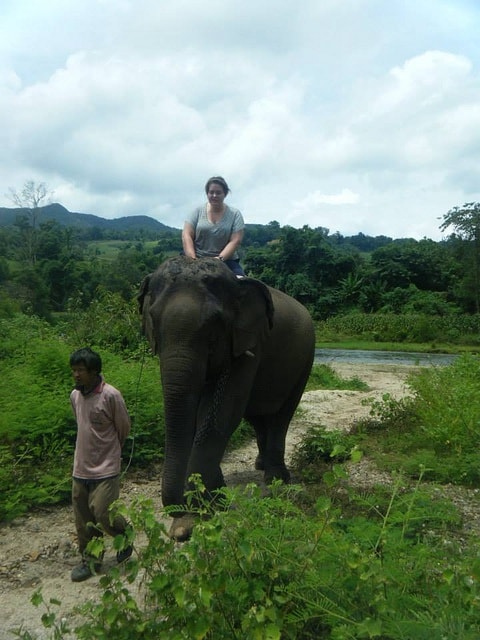}
\end{minipage}
\hspace{-1.5mm}
\begin{minipage}{0.1\linewidth}
\includegraphics[width=1\linewidth,  height=1\linewidth, scale=0.8, trim={0 0in 0 0in}, clip]{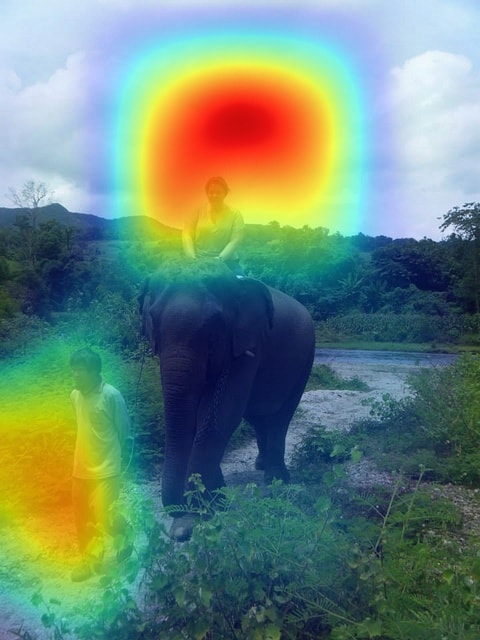}
\end{minipage}
\hspace{-1.5mm}
\begin{minipage}{0.1\linewidth}
\includegraphics[width=1\linewidth,  height=1\linewidth, scale=0.8, trim={0 0in 0 0in}, clip]{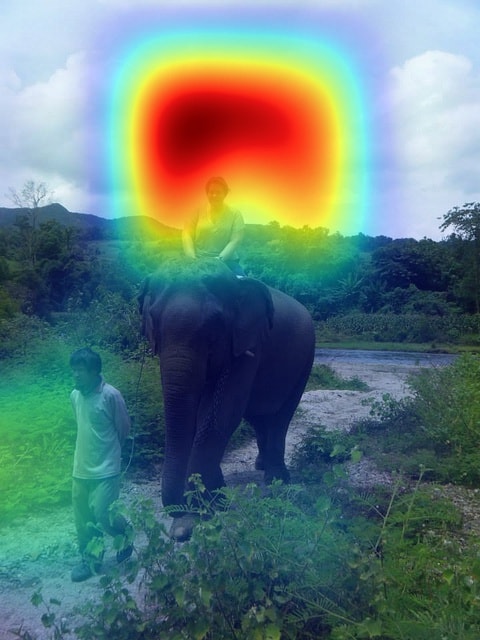}
\end{minipage}
\hspace{-1.5mm}
\begin{minipage}{0.1\linewidth}
\includegraphics[width=1\linewidth,  height=1\linewidth, scale=0.8, trim={0 0in 0 0in}, clip]{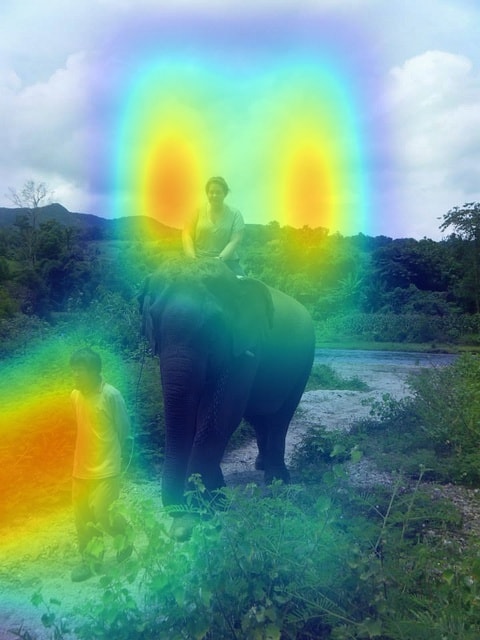}
\end{minipage}
\hspace{-1.5mm}
\begin{minipage}{0.1\linewidth}
\includegraphics[width=1\linewidth,  height=1\linewidth, scale=0.8, trim={0 0in 0 0in}, clip]{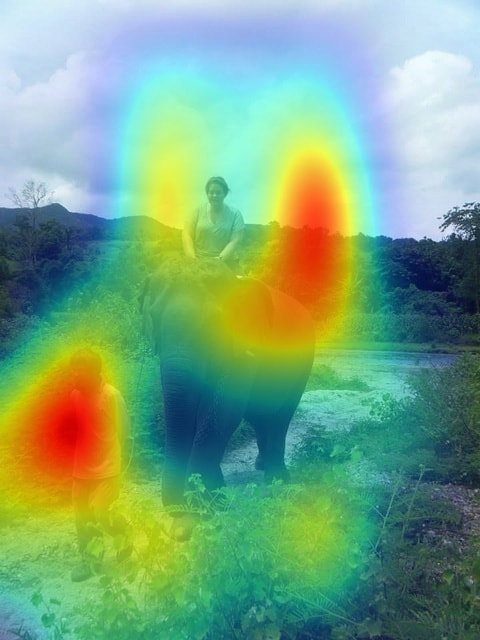}
\end{minipage}
\hspace{-1.5mm}
\begin{minipage}{0.1\linewidth}
\includegraphics[width=1\linewidth,  height=1\linewidth, scale=0.8, trim={0 0in 0 0in}, clip]{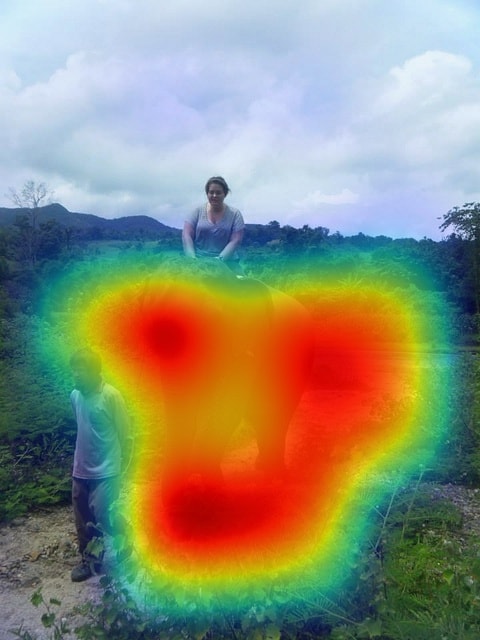}
\end{minipage}
\hspace{-1.5mm}
\begin{minipage}{0.1\linewidth}
\includegraphics[width=1\linewidth,  height=1\linewidth, scale=0.8, trim={0 0in 0 0in}, clip]{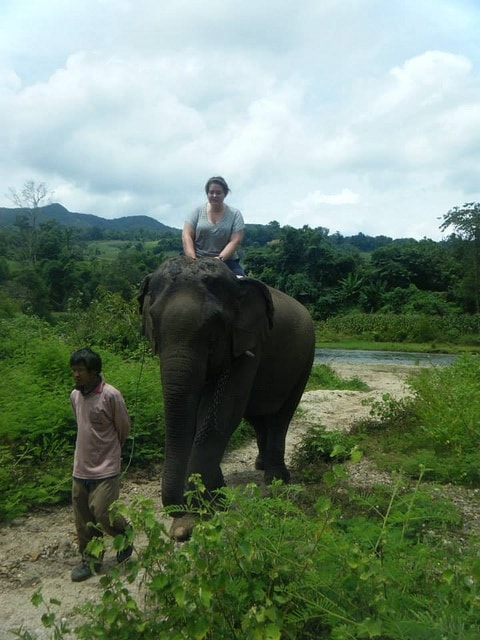}
\end{minipage}
\hspace{-1.5mm}
\begin{minipage}{0.1\linewidth}
\includegraphics[width=1\linewidth,  height=1\linewidth, scale=0.8, trim={0 0in 0 0in}, clip]{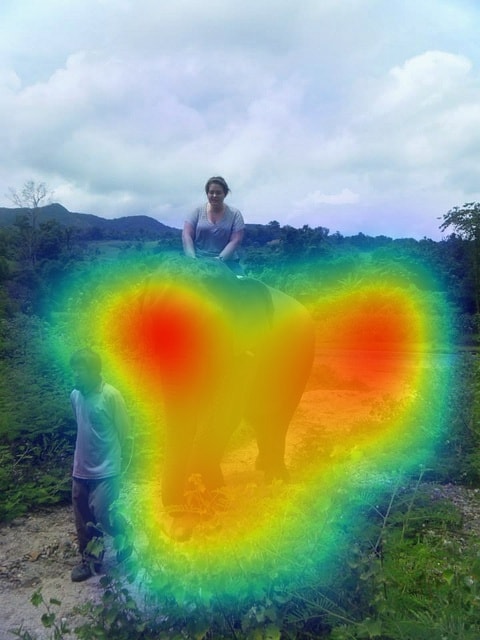}
\end{minipage}
\hspace{-1.5mm}
\begin{minipage}{0.1\linewidth}
\includegraphics[width=1\linewidth,  height=1\linewidth, scale=0.8, trim={0 0in 0 0in}, clip]{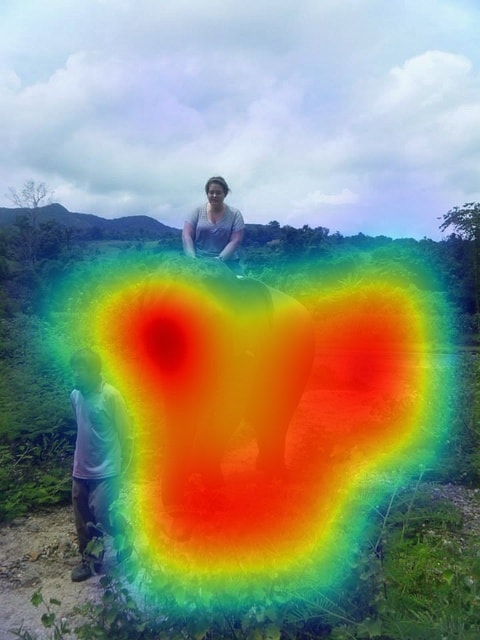}
\end{minipage}
\hspace{-1.5mm}
\\
\vspace{1mm}
\begin{minipage}{0.1\linewidth}
\begin{center}
\end{center}
\end{minipage}
\hspace{-1.5mm}
\begin{minipage}{0.1\linewidth}
\begin{center}
\tiny{a (0.87)}
\end{center}
\end{minipage}
\hspace{-1.5mm}
\begin{minipage}{0.1\linewidth}
\begin{center}
\tiny{man (0.91)}
\end{center}
\end{minipage}
\hspace{-1.5mm}
\begin{minipage}{0.1\linewidth}
\begin{center}
\tiny{riding (0.78)}
\end{center}
\end{minipage}
\hspace{-1.5mm}
\begin{minipage}{0.1\linewidth}
\begin{center}
\tiny{on (0.86)}
\end{center}
\end{minipage}
\hspace{-1.5mm}
\begin{minipage}{0.1\linewidth}
\begin{center}
\tiny{top (0.89)}
\end{center}
\end{minipage}
\hspace{-1.5mm}
\begin{minipage}{0.1\linewidth}
\begin{center}
\tiny{of (0.35)}
\end{center}
\end{minipage}
\hspace{-1.5mm}
\begin{minipage}{0.1\linewidth}
\begin{center}
\tiny{an (0.84)}
\end{center}
\end{minipage}
\hspace{-1.5mm}
\begin{minipage}{0.1\linewidth}
\begin{center}
\tiny{elephant (0.88)}
\end{center}
\end{minipage}
\hspace{-1.5mm}
\\
\begin{minipage}{0.1\linewidth}
\includegraphics[width=1\linewidth,  height=1\linewidth, scale=0.8, trim={0 0in 0 0in}, clip]{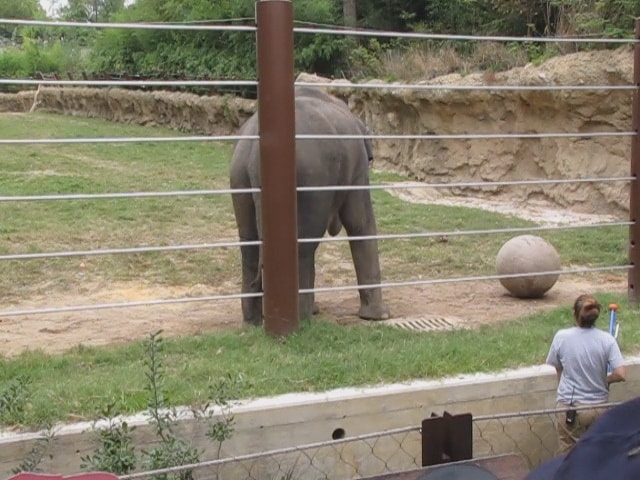}
\end{minipage}
\hspace{-1.5mm}
\begin{minipage}{0.1\linewidth}
\includegraphics[width=1\linewidth,  height=1\linewidth, scale=0.8, trim={0 0in 0 0in}, clip]{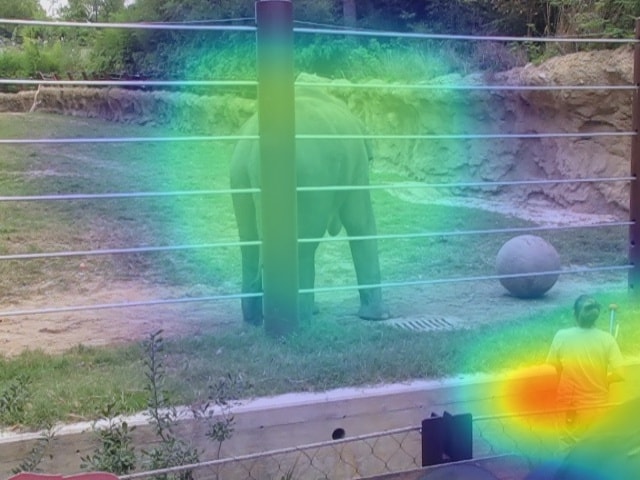}
\end{minipage}
\hspace{-1.5mm}
\begin{minipage}{0.1\linewidth}
\includegraphics[width=1\linewidth,  height=1\linewidth, scale=0.8, trim={0 0in 0 0in}, clip]{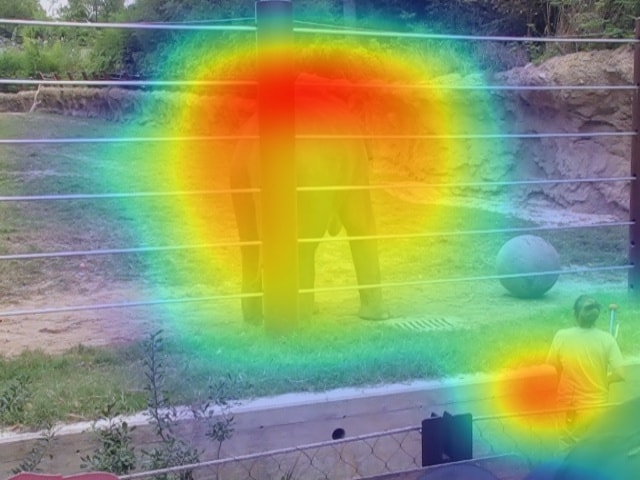}
\end{minipage}
\hspace{-1.5mm}
\begin{minipage}{0.1\linewidth}
\includegraphics[width=1\linewidth,  height=1\linewidth, scale=0.8, trim={0 0in 0 0in}, clip]{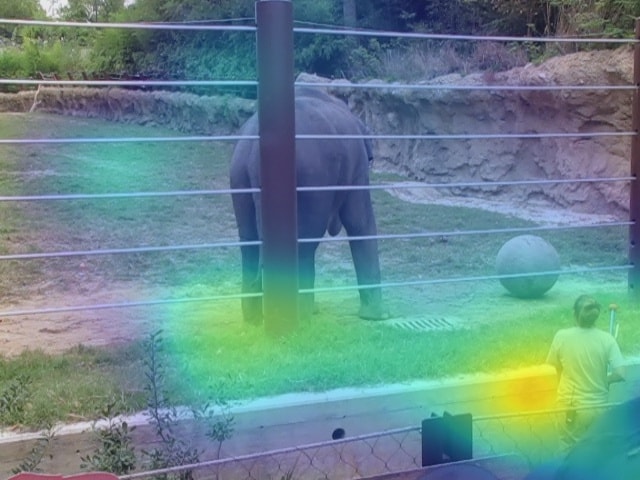}
\end{minipage}
\hspace{-1.5mm}
\begin{minipage}{0.1\linewidth}
\includegraphics[width=1\linewidth,  height=1\linewidth, scale=0.8, trim={0 0in 0 0in}, clip]{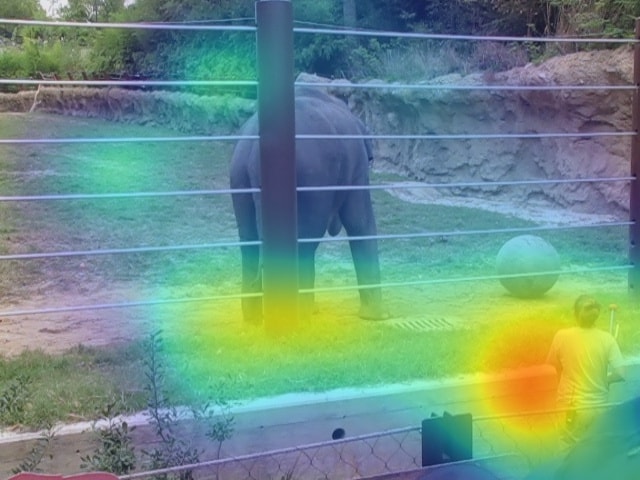}
\end{minipage}
\hspace{-1.5mm}
\begin{minipage}{0.1\linewidth}
\includegraphics[width=1\linewidth,  height=1\linewidth, scale=0.8, trim={0 0in 0 0in}, clip]{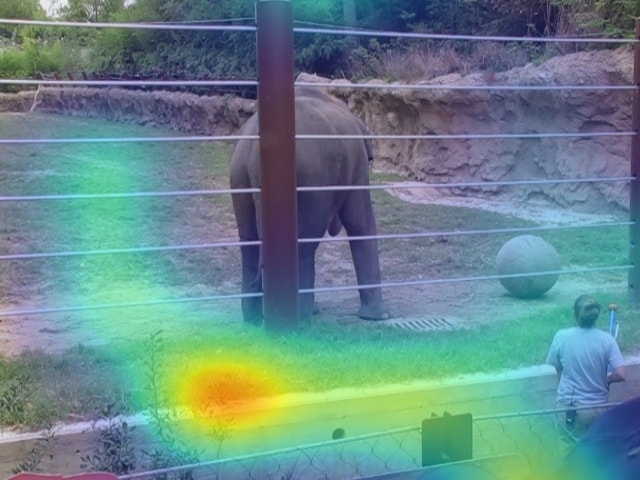}
\end{minipage}
\hspace{-1.5mm}
\begin{minipage}{0.1\linewidth}
\includegraphics[width=1\linewidth,  height=1\linewidth, scale=0.8, trim={0 0in 0 0in}, clip]{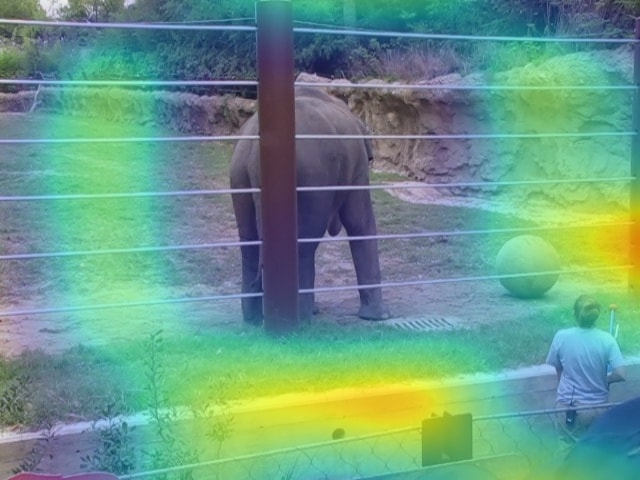}
\end{minipage}
\hspace{-1.5mm}
\begin{minipage}{0.1\linewidth}
\includegraphics[width=1\linewidth,  height=1\linewidth, scale=0.8, trim={0 0in 0 0in}, clip]{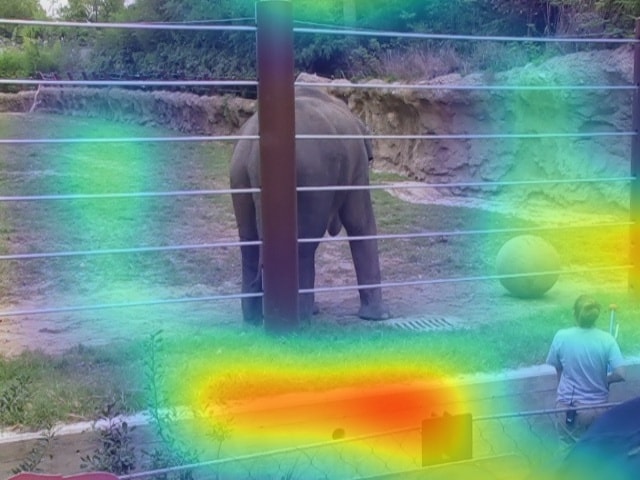}
\end{minipage}
\hspace{-1.5mm}
\begin{minipage}{0.1\linewidth}
\includegraphics[width=1\linewidth,  height=1\linewidth, scale=0.8, trim={0 0in 0 0in}, clip]{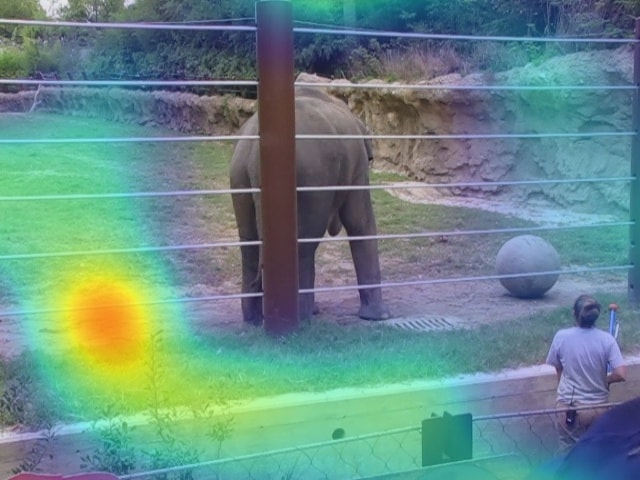}
\end{minipage}
\hspace{-1.5mm}
\\
\vspace{1mm}
\begin{minipage}{0.1\linewidth}
\begin{center}
\end{center}
\end{minipage}
\hspace{-1.5mm}
\begin{minipage}{0.1\linewidth}
\begin{center}
\tiny{an (0.79)}
\end{center}
\end{minipage}
\hspace{-1.5mm}
\begin{minipage}{0.1\linewidth}
\begin{center}
\tiny{elephant (0.87)}
\end{center}
\end{minipage}
\hspace{-1.5mm}
\begin{minipage}{0.1\linewidth}
\begin{center}
\tiny{standing (0.67)}
\end{center}
\end{minipage}
\hspace{-1.5mm}
\begin{minipage}{0.1\linewidth}
\begin{center}
\tiny{in (0.78)}
\end{center}
\end{minipage}
\hspace{-1.5mm}
\begin{minipage}{0.1\linewidth}
\begin{center}
\tiny{a (0.75)}
\end{center}
\end{minipage}
\hspace{-1.5mm}
\begin{minipage}{0.1\linewidth}
\begin{center}
\tiny{fenced (0.74)}
\end{center}
\end{minipage}
\hspace{-1.5mm}
\begin{minipage}{0.1\linewidth}
\begin{center}
\tiny{in (0.83)}
\end{center}
\end{minipage}
\hspace{-1.5mm}
\begin{minipage}{0.1\linewidth}
\begin{center}
\tiny{area (0.77)}
\end{center}
\end{minipage}
\hspace{-1.5mm}
\\
\begin{minipage}{0.1\linewidth}
\includegraphics[width=1\linewidth,  height=1\linewidth, scale=0.8, trim={0 0in 0 0in}, clip]{supl2/img_55.jpg}
\end{minipage}
\hspace{-1.5mm}
\begin{minipage}{0.1\linewidth}
\includegraphics[width=1\linewidth,  height=1\linewidth, scale=0.8, trim={0 0in 0 0in}, clip]{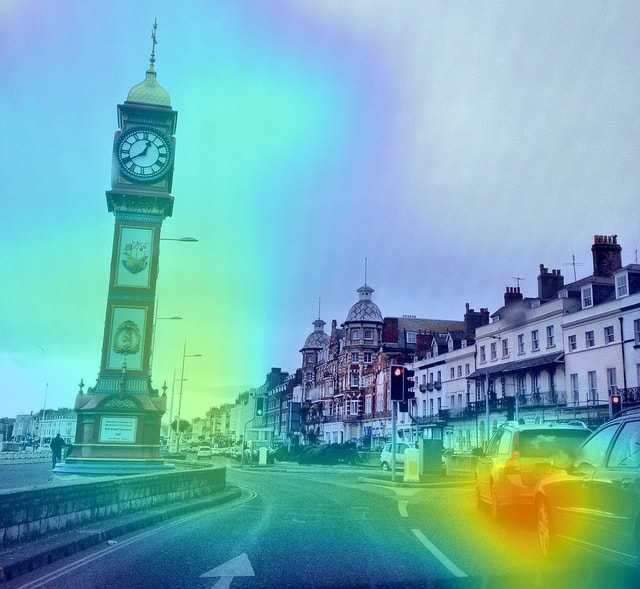}
\end{minipage}
\hspace{-1.5mm}
\begin{minipage}{0.1\linewidth}
\includegraphics[width=1\linewidth,  height=1\linewidth, scale=0.8, trim={0 0in 0 0in}, clip]{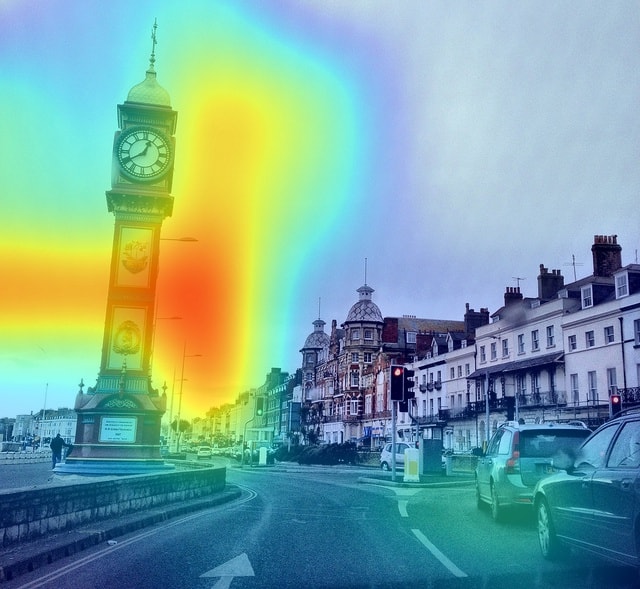}
\end{minipage}
\hspace{-1.5mm}
\begin{minipage}{0.1\linewidth}
\includegraphics[width=1\linewidth,  height=1\linewidth, scale=0.8, trim={0 0in 0 0in}, clip]{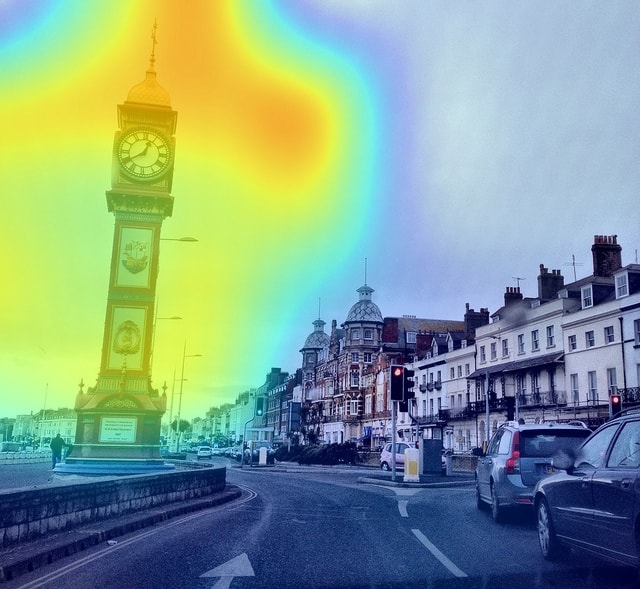}
\end{minipage}
\hspace{-1.5mm}
\begin{minipage}{0.1\linewidth}
\includegraphics[width=1\linewidth,  height=1\linewidth, scale=0.8, trim={0 0in 0 0in}, clip]{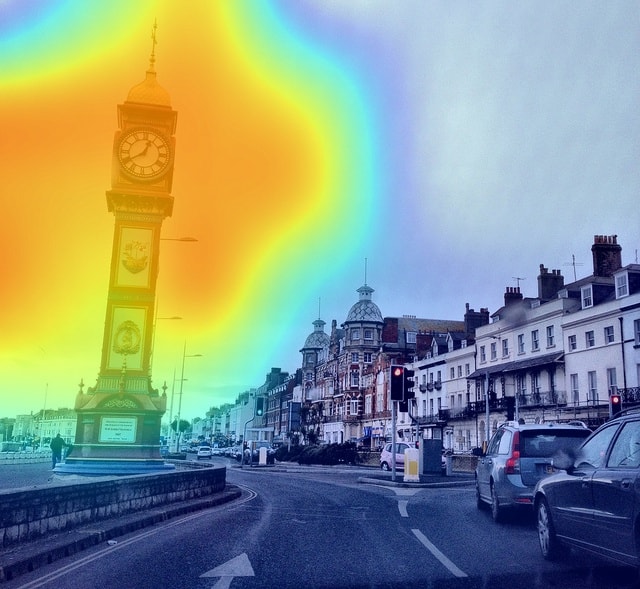}
\end{minipage}
\hspace{-1.5mm}
\begin{minipage}{0.1\linewidth}
\includegraphics[width=1\linewidth,  height=1\linewidth, scale=0.8, trim={0 0in 0 0in}, clip]{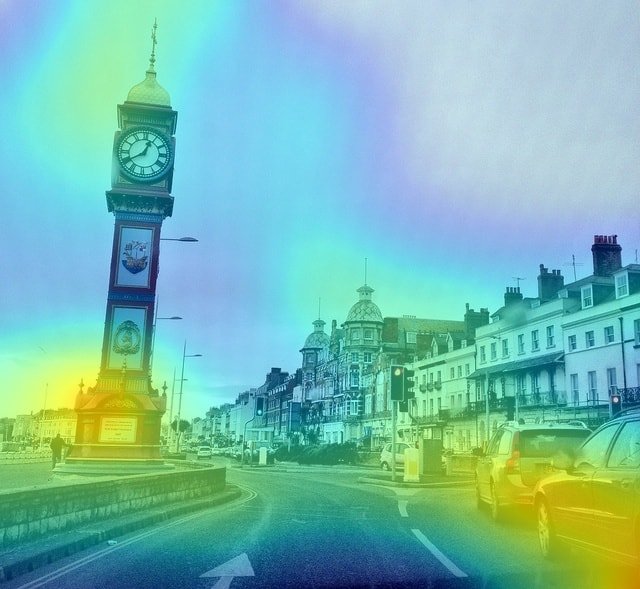}
\end{minipage}
\hspace{-1.5mm}
\begin{minipage}{0.1\linewidth}
\includegraphics[width=1\linewidth,  height=1\linewidth, scale=0.8, trim={0 0in 0 0in}, clip]{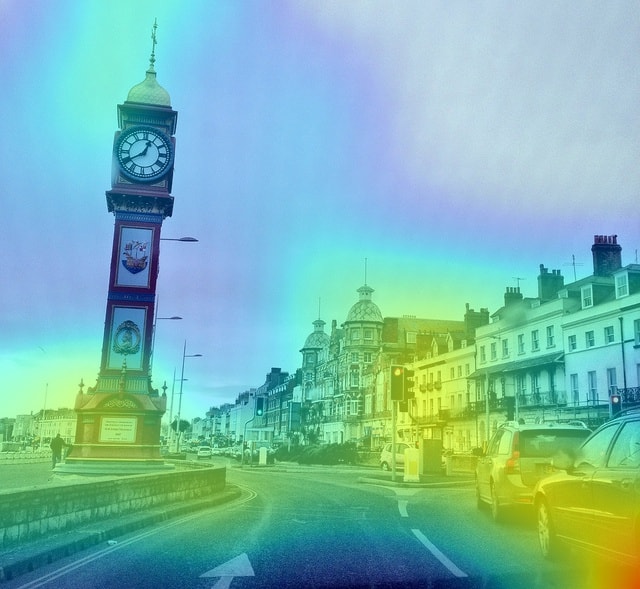}
\end{minipage}
\hspace{-1.5mm}
\begin{minipage}{0.1\linewidth}
\includegraphics[width=1\linewidth,  height=1\linewidth, scale=0.8, trim={0 0in 0 0in}, clip]{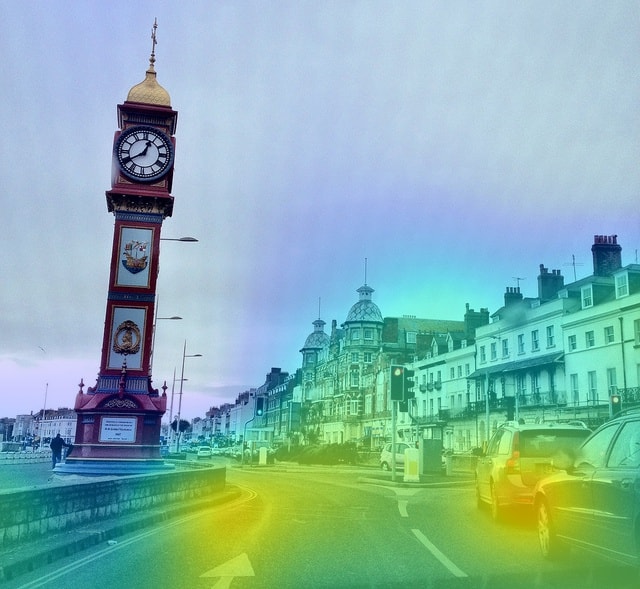}
\end{minipage}
\hspace{-1.5mm}
\begin{minipage}{0.1\linewidth}
\includegraphics[width=1\linewidth,  height=1\linewidth, scale=0.8, trim={0 0in 0 0in}, clip]{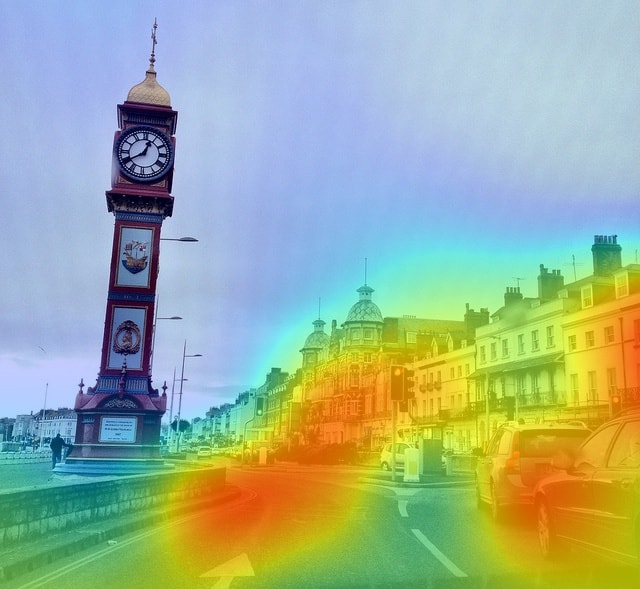}
\end{minipage}
\hspace{-1.5mm}
\\
\vspace{1mm}
\begin{minipage}{0.1\linewidth}
\begin{center}
\end{center}
\end{minipage}
\hspace{-1.5mm}
\begin{minipage}{0.1\linewidth}
\begin{center}
\tiny{a (0.76)}
\end{center}
\end{minipage}
\hspace{-1.5mm}
\begin{minipage}{0.1\linewidth}
\begin{center}
\tiny{tall (0.83)}
\end{center}
\end{minipage}
\hspace{-1.5mm}
\begin{minipage}{0.1\linewidth}
\begin{center}
\tiny{clock (0.73)}
\end{center}
\end{minipage}
\hspace{-1.5mm}
\begin{minipage}{0.1\linewidth}
\begin{center}
\tiny{tower (0.77)}
\end{center}
\end{minipage}
\hspace{-1.5mm}
\begin{minipage}{0.1\linewidth}
\begin{center}
\tiny{towering (0.72)}
\end{center}
\end{minipage}
\hspace{-1.5mm}
\begin{minipage}{0.1\linewidth}
\begin{center}
\tiny{over (0.79)}
\end{center}
\end{minipage}
\hspace{-1.5mm}
\begin{minipage}{0.1\linewidth}
\begin{center}
\tiny{a (0.68)}
\end{center}
\end{minipage}
\hspace{-1.5mm}
\begin{minipage}{0.1\linewidth}
\begin{center}
\tiny{city (0.82)}
\end{center}
\end{minipage}
\hspace{-1.5mm}
\captionof{figure}{\small{Example of generated caption, spatial attention and visual grounding probability.}}
\label{vis2}
\vspace{-3mm}
\end{figure*}

Fig~\ref{vis1} and Fig~\ref{vis2} show additional visualization of spatial and temporal attention. 

\subsection{Visualization of Weakly Supervised Localization}

\begin{figure*}[tb]
\center
\begin{minipage}{0.16\linewidth}
\includegraphics[width=1\linewidth,  height=1\linewidth, scale=0.8, trim={0 0in 0 0in}, clip]{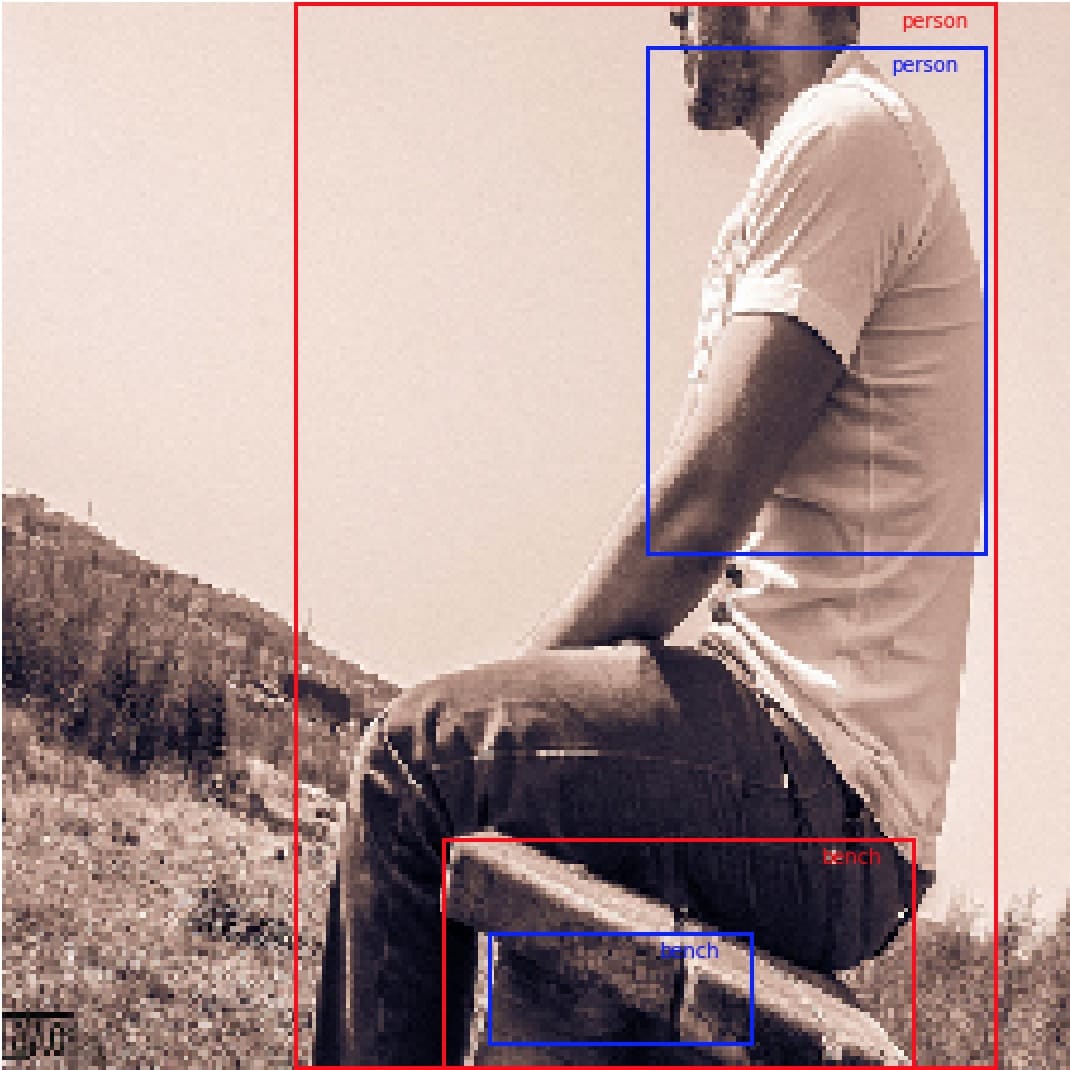}
\end{minipage}
\hspace{0.5mm}
\begin{minipage}{0.16\linewidth}
\includegraphics[width=1\linewidth,  height=1\linewidth, scale=0.8, trim={0 0in 0 0in}, clip]{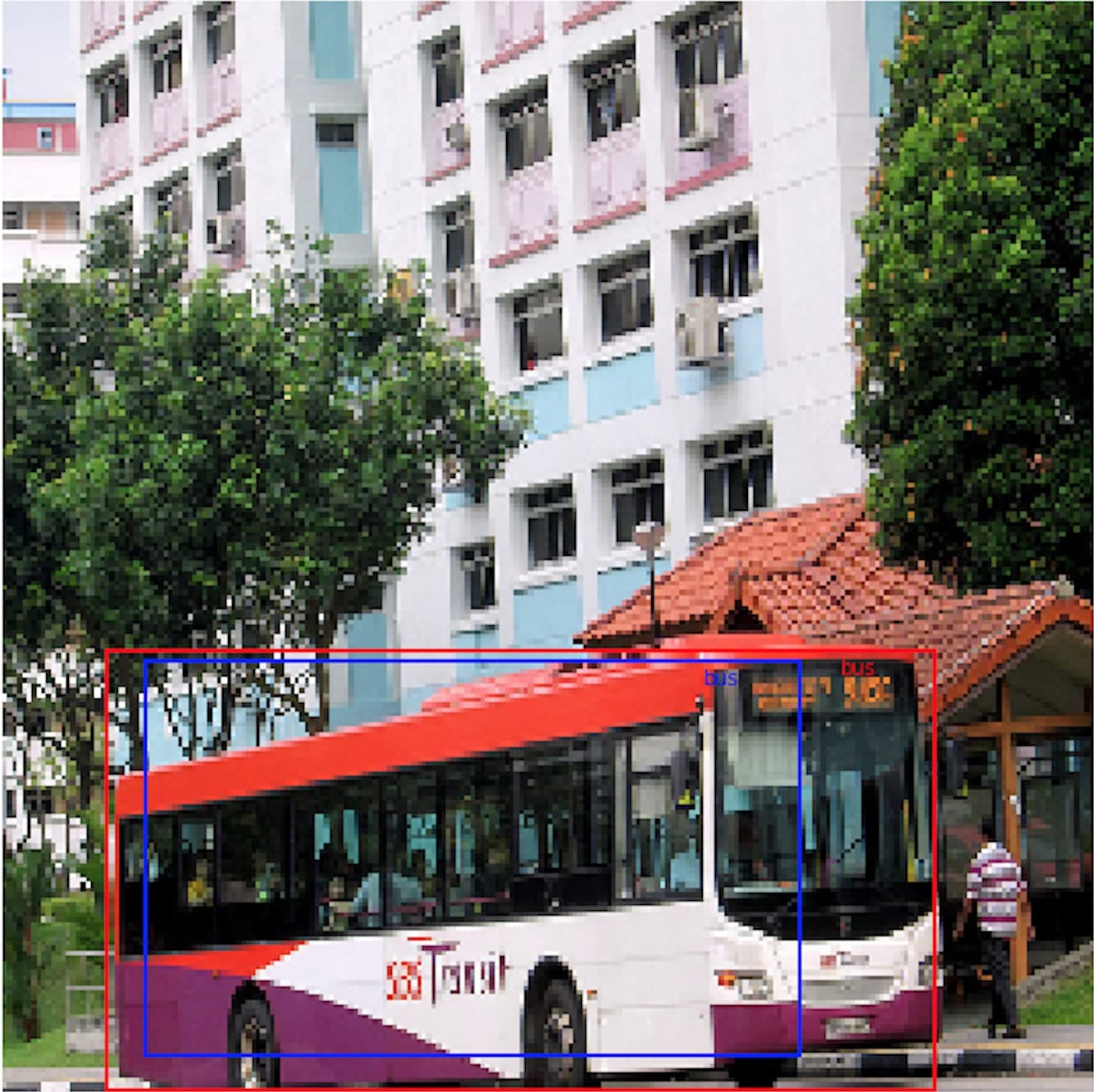}
\end{minipage}
\hspace{0.5mm}
\begin{minipage}{0.16\linewidth}
\includegraphics[width=1\linewidth,  height=1\linewidth, scale=0.8, trim={0 0in 0 0in}, clip]{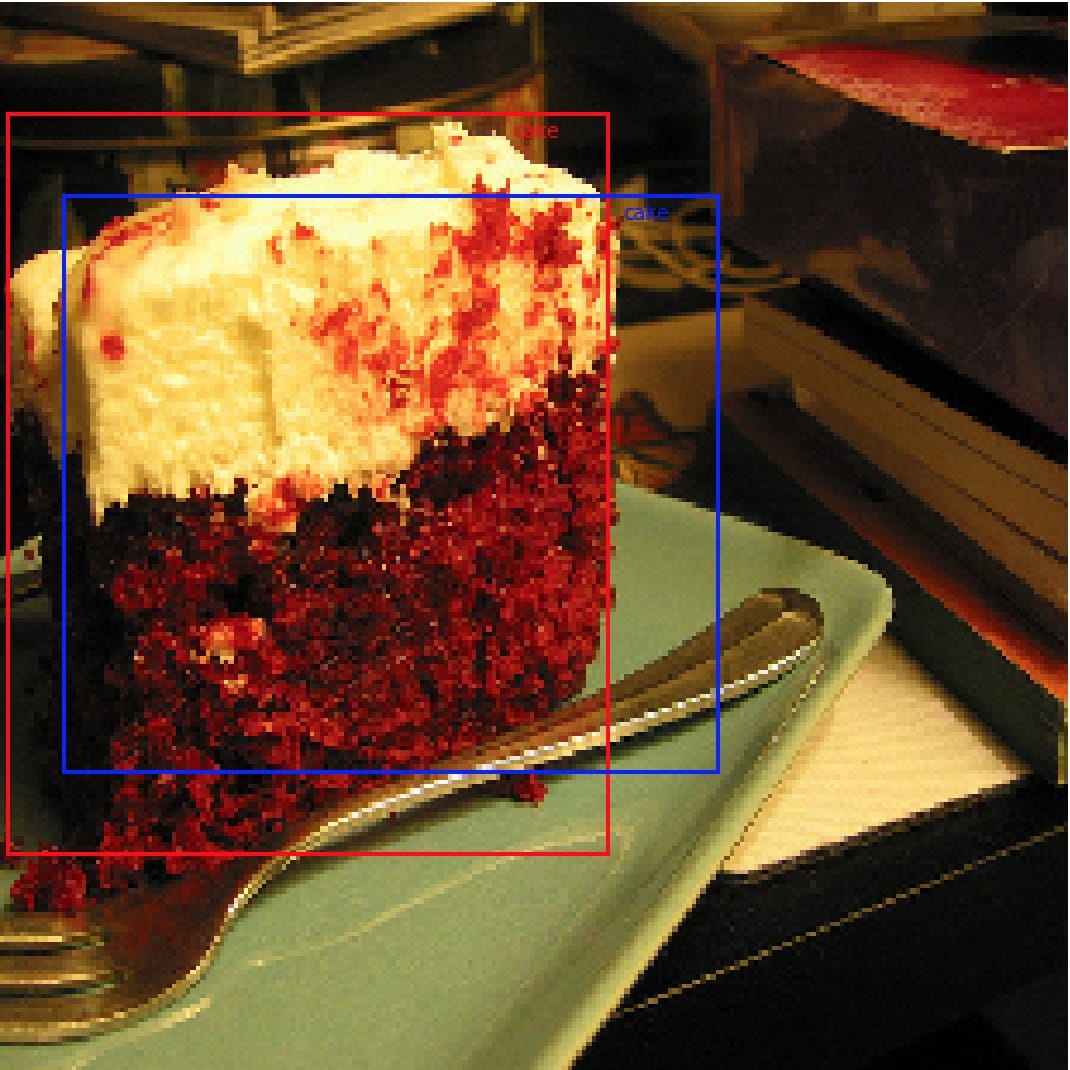}
\end{minipage}
\hspace{0.5mm}
\begin{minipage}{0.16\linewidth}
\includegraphics[width=1\linewidth,  height=1\linewidth, scale=0.8, trim={0 0in 0 0in}, clip]{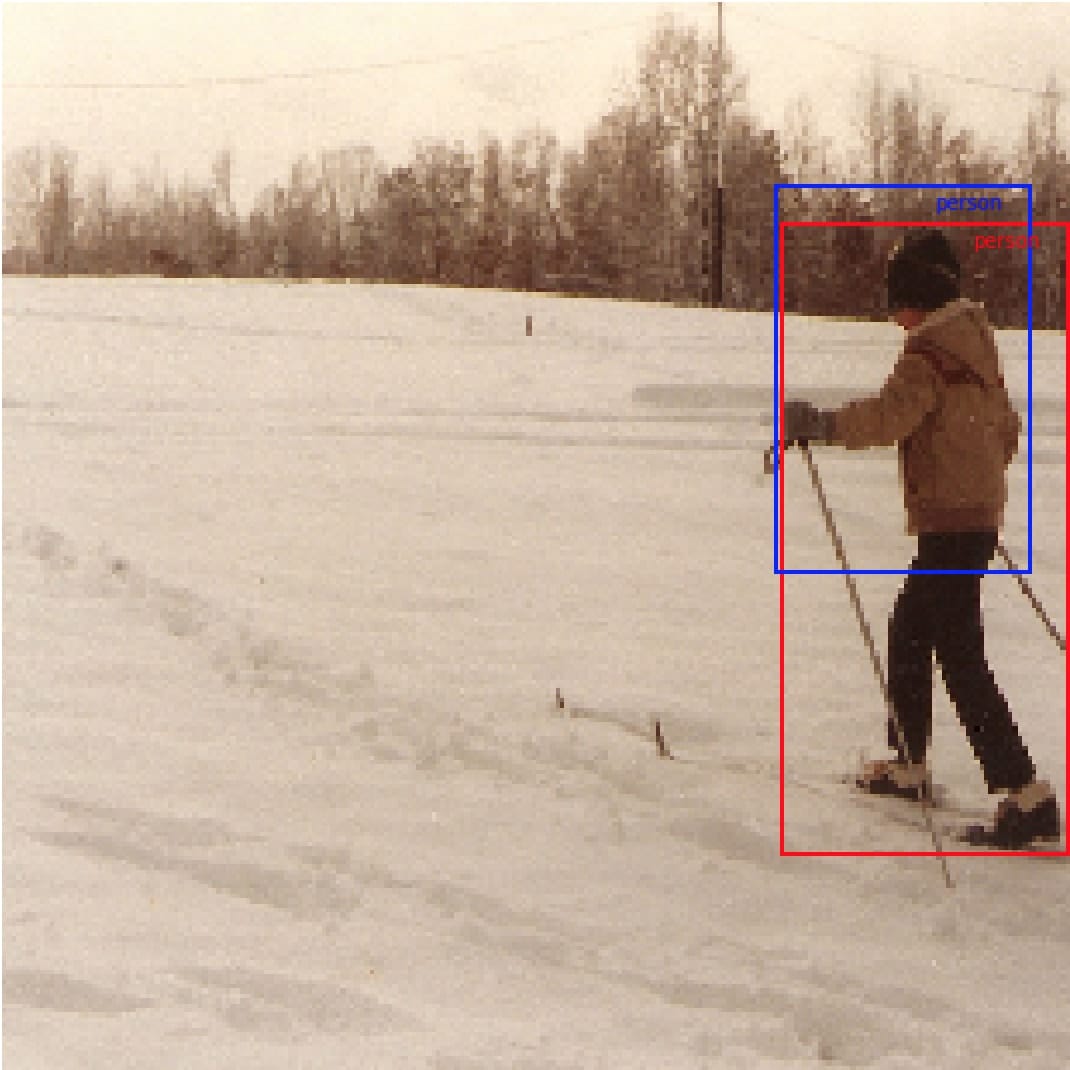}
\end{minipage}
\hspace{0.5mm}
\begin{minipage}{0.16\linewidth}
\includegraphics[width=1\linewidth,  height=1\linewidth, scale=0.8, trim={0 0in 0 0in}, clip]{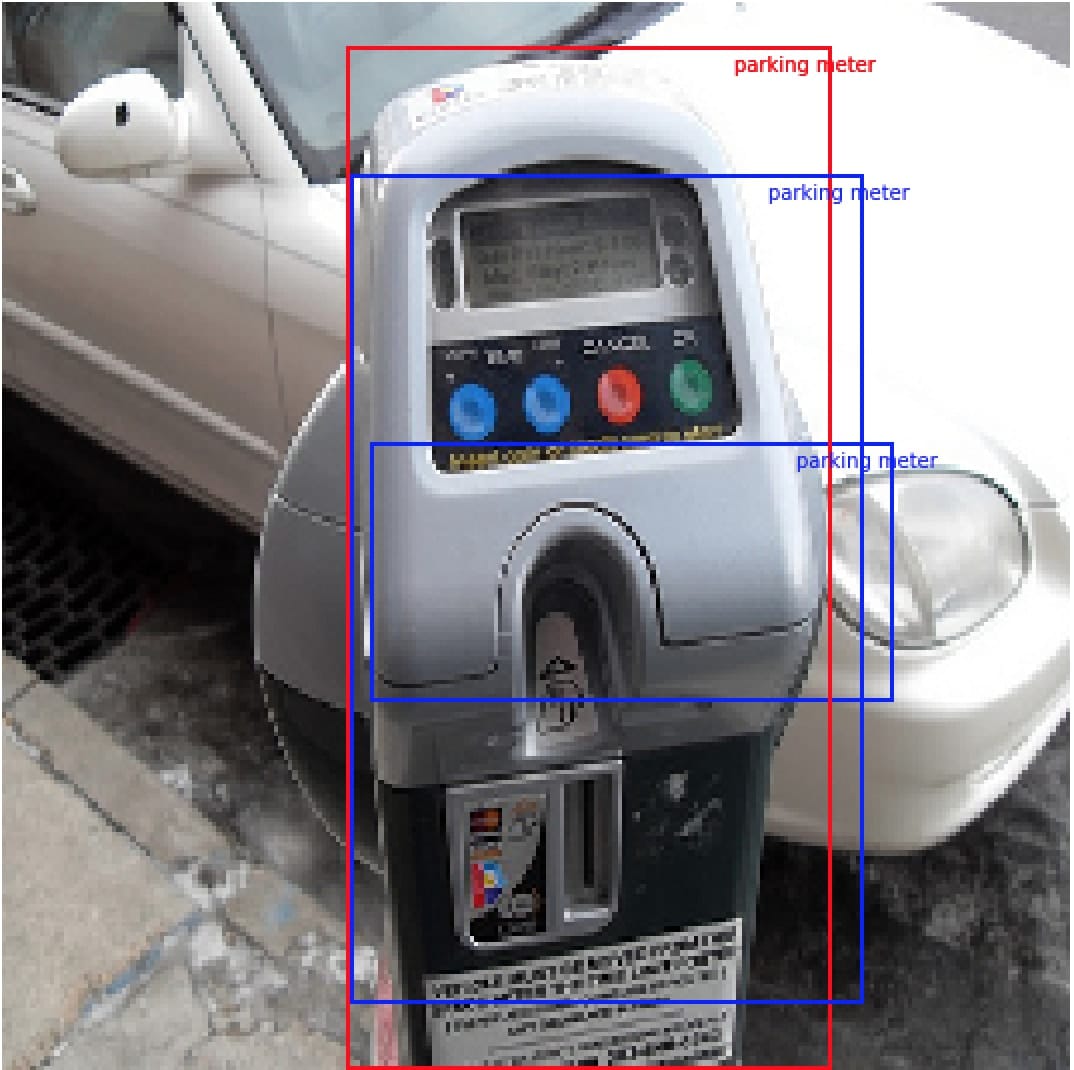}
\end{minipage}
\begin{minipage}{0.16\linewidth}
\begin{center}
\small{a man sitting on top of a wooden bench. }
\end{center}
\end{minipage}
\hspace{0.5mm}
\begin{minipage}{0.16\linewidth}
\begin{center}
\small{a red and white us parked in front of a building. }
\end{center}
\end{minipage}
\hspace{0.5mm}
\begin{minipage}{0.16\linewidth}
\begin{center}
\small{a piece of cake sitting on top of a white plate.}
\end{center}
\end{minipage}
\hspace{0.5mm}
\begin{minipage}{0.16\linewidth}
\begin{center}
\small{a man riding skis down a snow covered slope.}
\end{center}
\end{minipage}
\hspace{0.5mm}
\begin{minipage}{0.16\linewidth}
\begin{center}
\small{a parking meter sitting on the side of a road.}
\end{center}
\end{minipage}
\\
\vspace{1mm}
\begin{minipage}{0.16\linewidth}
\includegraphics[width=1\linewidth,  height=1\linewidth, scale=0.8, trim={0 0in 0 0in}, clip]{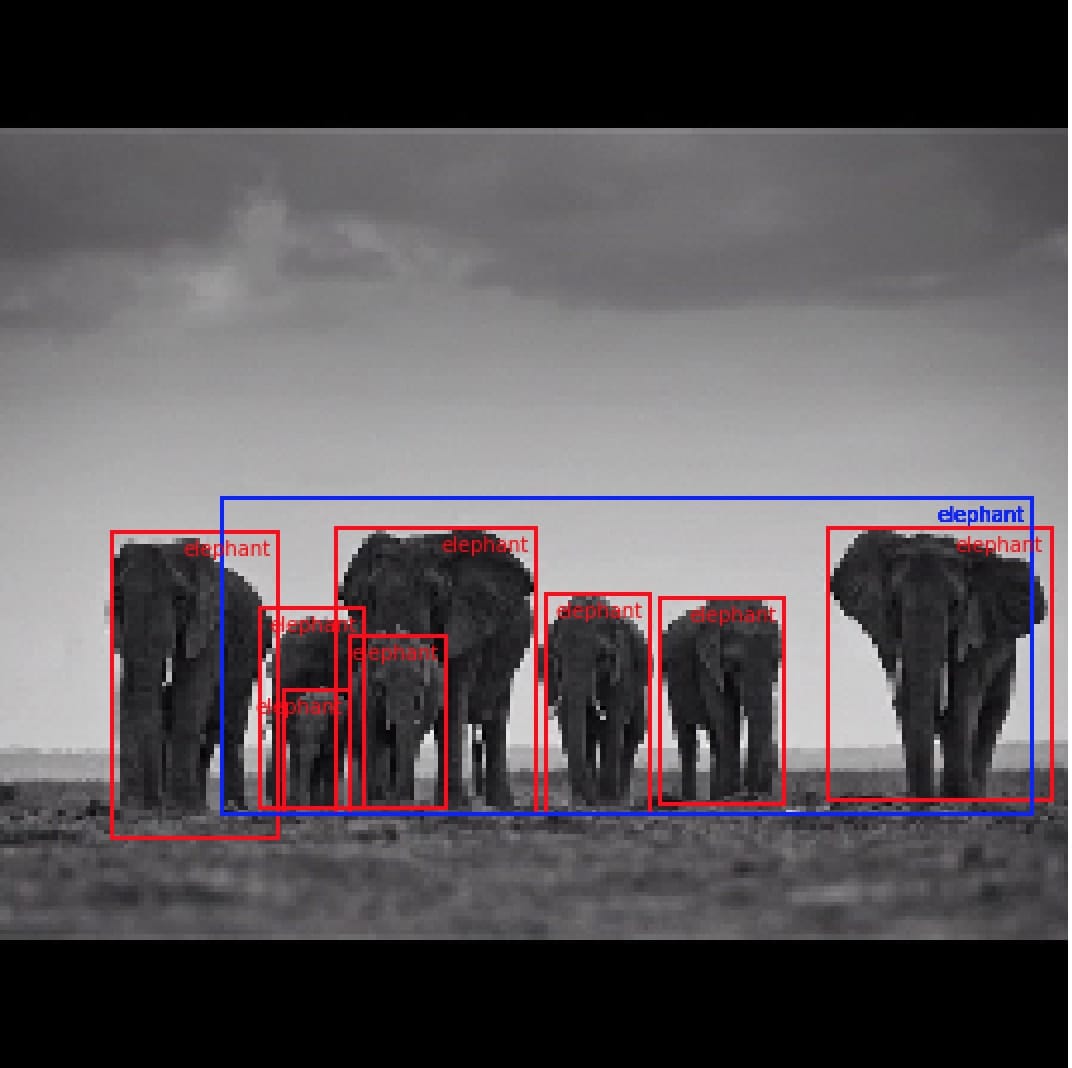}
\end{minipage}
\hspace{0.5mm}
\begin{minipage}{0.16\linewidth}
\includegraphics[width=1\linewidth,  height=1\linewidth, scale=0.8, trim={0 0in 0 0in}, clip]{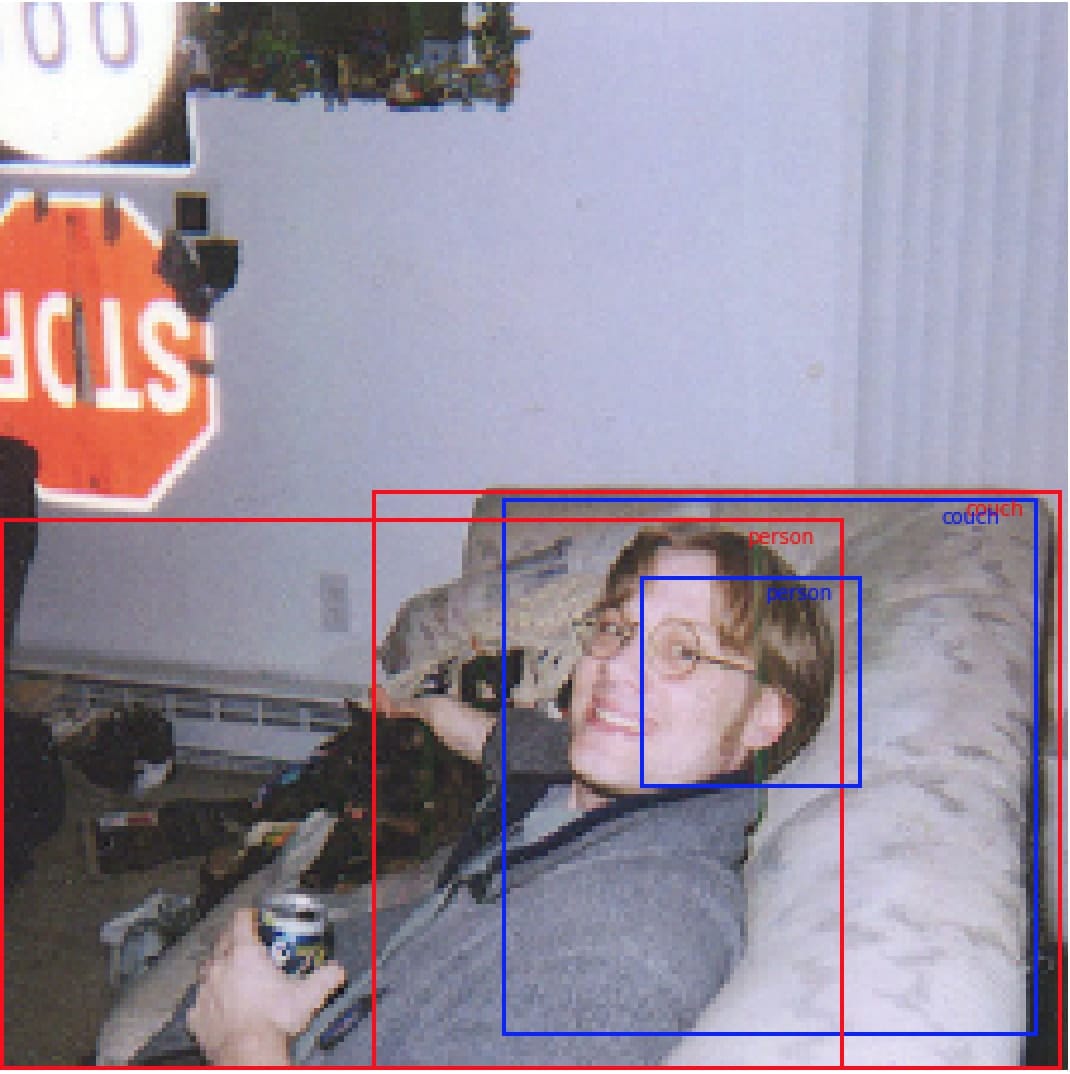}
\end{minipage}
\hspace{0.5mm}
\begin{minipage}{0.16\linewidth}
\includegraphics[width=1\linewidth,  height=1\linewidth, scale=0.8, trim={0 0in 0 0in}, clip]{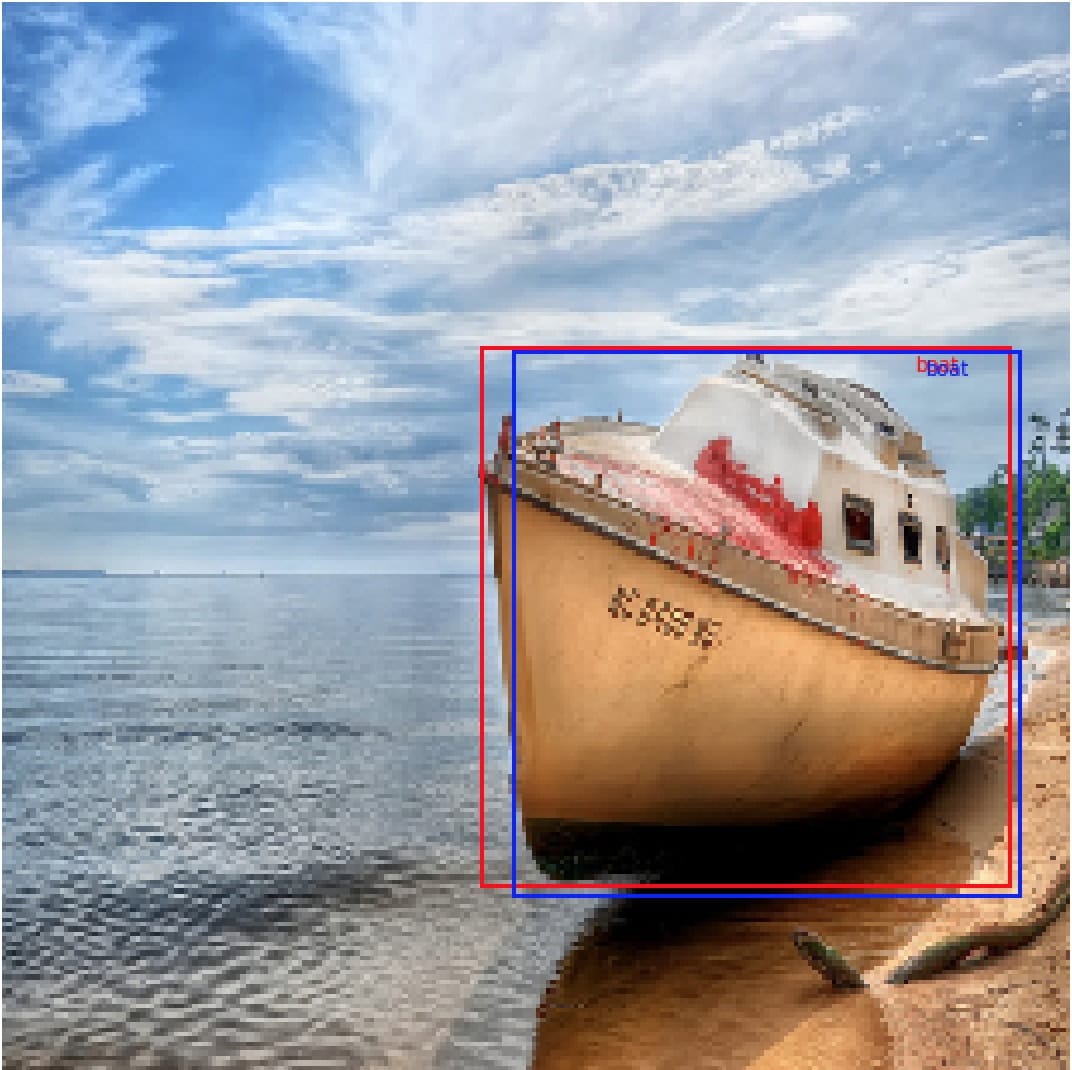}
\end{minipage}
\hspace{0.5mm}
\begin{minipage}{0.16\linewidth}
\includegraphics[width=1\linewidth,  height=1\linewidth, scale=0.8, trim={0 0in 0 0in}, clip]{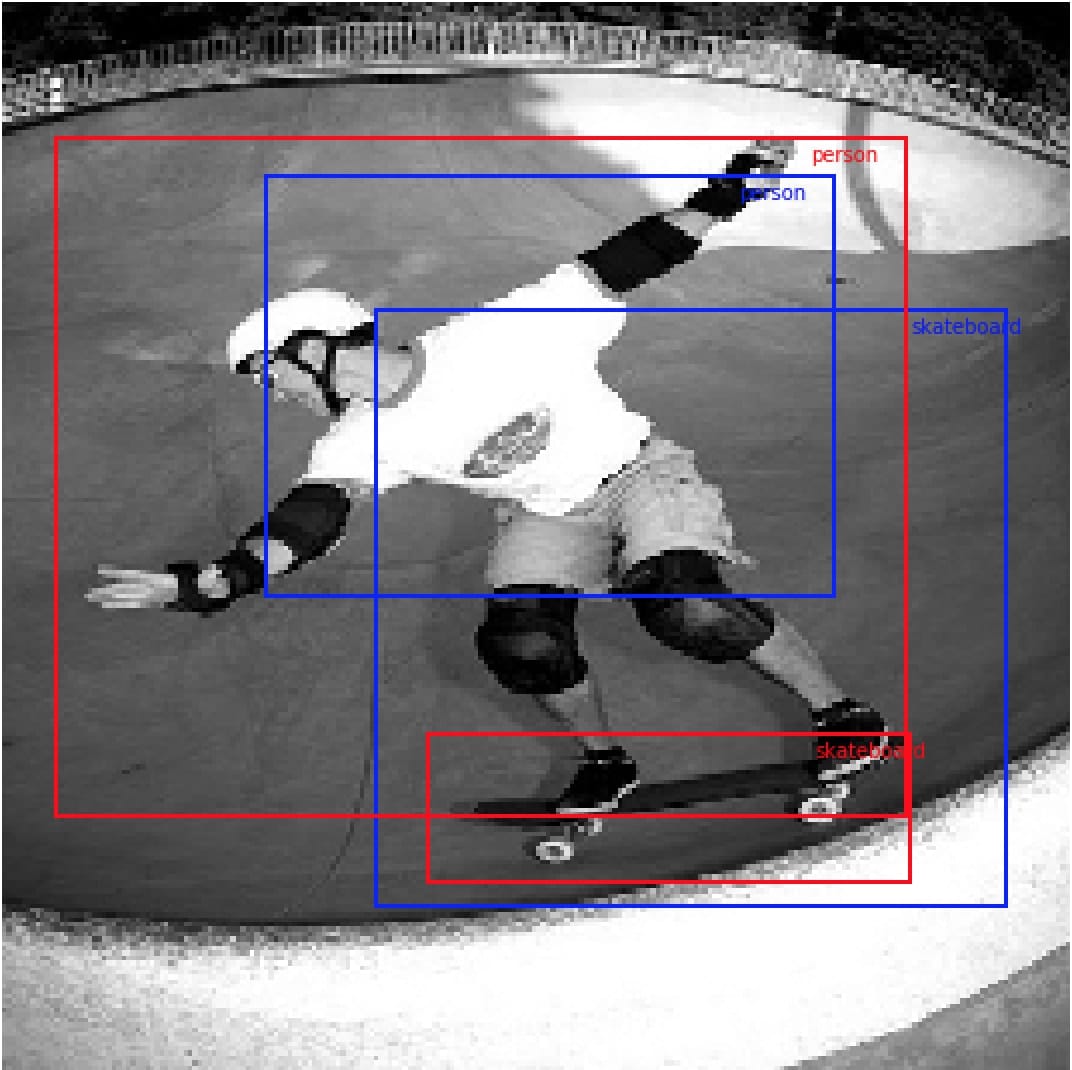}
\end{minipage}
\hspace{0.5mm}
\begin{minipage}{0.16\linewidth}
\includegraphics[width=1\linewidth,  height=1\linewidth, scale=0.8, trim={0 0in 0 0in}, clip]{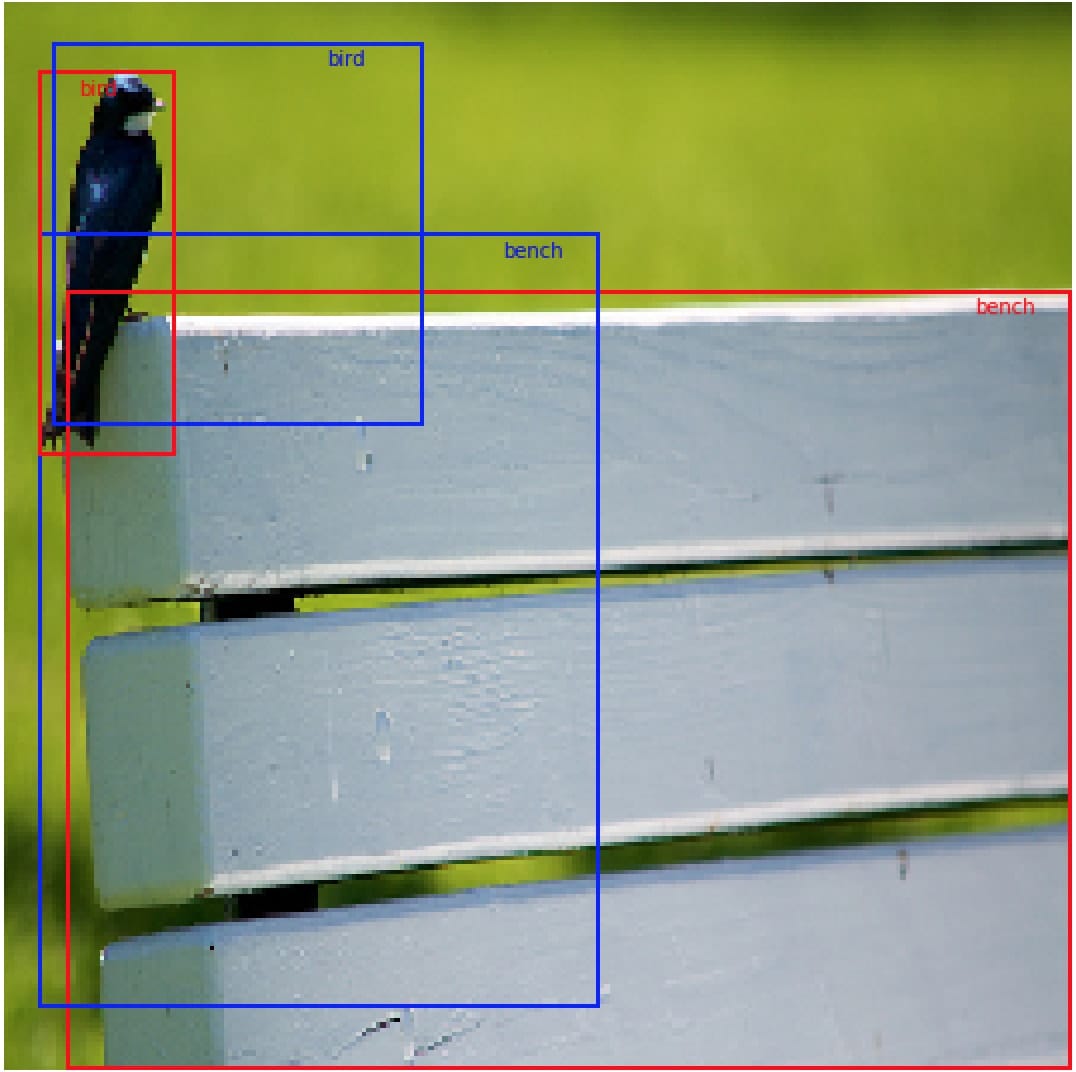}
\end{minipage}
\begin{minipage}{0.16\linewidth}
\begin{center}
\small{a herd of elephants standing next to each other.}
\end{center}
\end{minipage}
\hspace{0.5mm}
\begin{minipage}{0.16\linewidth}
\begin{center}
\small{a man laying on a couch holding a remote control.}
\end{center}
\end{minipage}
\hspace{0.5mm}
\begin{minipage}{0.16\linewidth}
\begin{center}
\small{a boat sitting on top of a sandy beach.}
\end{center}
\end{minipage}
\hspace{0.5mm}
\begin{minipage}{0.16\linewidth}
\begin{center}
\small{a man riding a skateboard up the side of a ramp.}
\end{center}
\end{minipage}
\hspace{0.5mm}
\begin{minipage}{0.16\linewidth}
\begin{center}
\small{a black bird sitting on top of a wooden bench.}
\end{center}
\end{minipage}
\captionof{figure}{Visualization of generated captions and weakly supervised localization result. Red bounding box is the ground truth annotation, blue bounding box is the predicted location using spatial attention map.}
\vspace{-3mm}
\label{vis3}
\end{figure*}

Fig~\ref{vis3} shows the visualization of  weakly supervised localization.  


\end{document}